%% file: main.tex
\documentclass[letterpaper]{article} 
\usepackage{aaai2026}  
\nocopyright
\usepackage{times}  
\usepackage{helvet}  
\usepackage{courier}  
\usepackage[hyphens]{url}  
\usepackage{graphicx} 
\usepackage{natbib}  
\usepackage{caption} 
\usepackage{algorithm}
\usepackage{algorithmic}

\usepackage{newfloat}
\usepackage{listings}
\DeclareCaptionStyle{ruled}{labelfont=normalfont,labelsep=colon,strut=off} 
\floatstyle{ruled}
\newfloat{listing}{tb}{lst}{}
\floatname{listing}{Listing}
\usepackage{amsmath}
\usepackage{amssymb}
\usepackage{booktabs}
\usepackage{multirow}
\usepackage{lipsum}
\usepackage{mathtools}
\usepackage{hyperref}
\usepackage[capitalize]{cleveref}

\title{CoherenDream: Boosting Holistic Text Coherence in 3D Generation via Multimodal Large Language Models Feedback}
\author{
    Chenhan Jiang\textsuperscript{\rm 1}\thanks{Corresponding author: jchcyan@gmail.com},
    Yihan Zeng\textsuperscript{\rm 2},
    Dit-Yan Yeung\textsuperscript{\rm 1}
}
\affiliations{
    \textsuperscript{\rm 1}Hong Kong University of Science and Technology\\
    \textsuperscript{\rm 2}Shanghai Jiao Tong University
}

\begin{document}

\maketitle

\begin{figure*}[t]
  \centering
  \includegraphics[width=\textwidth]{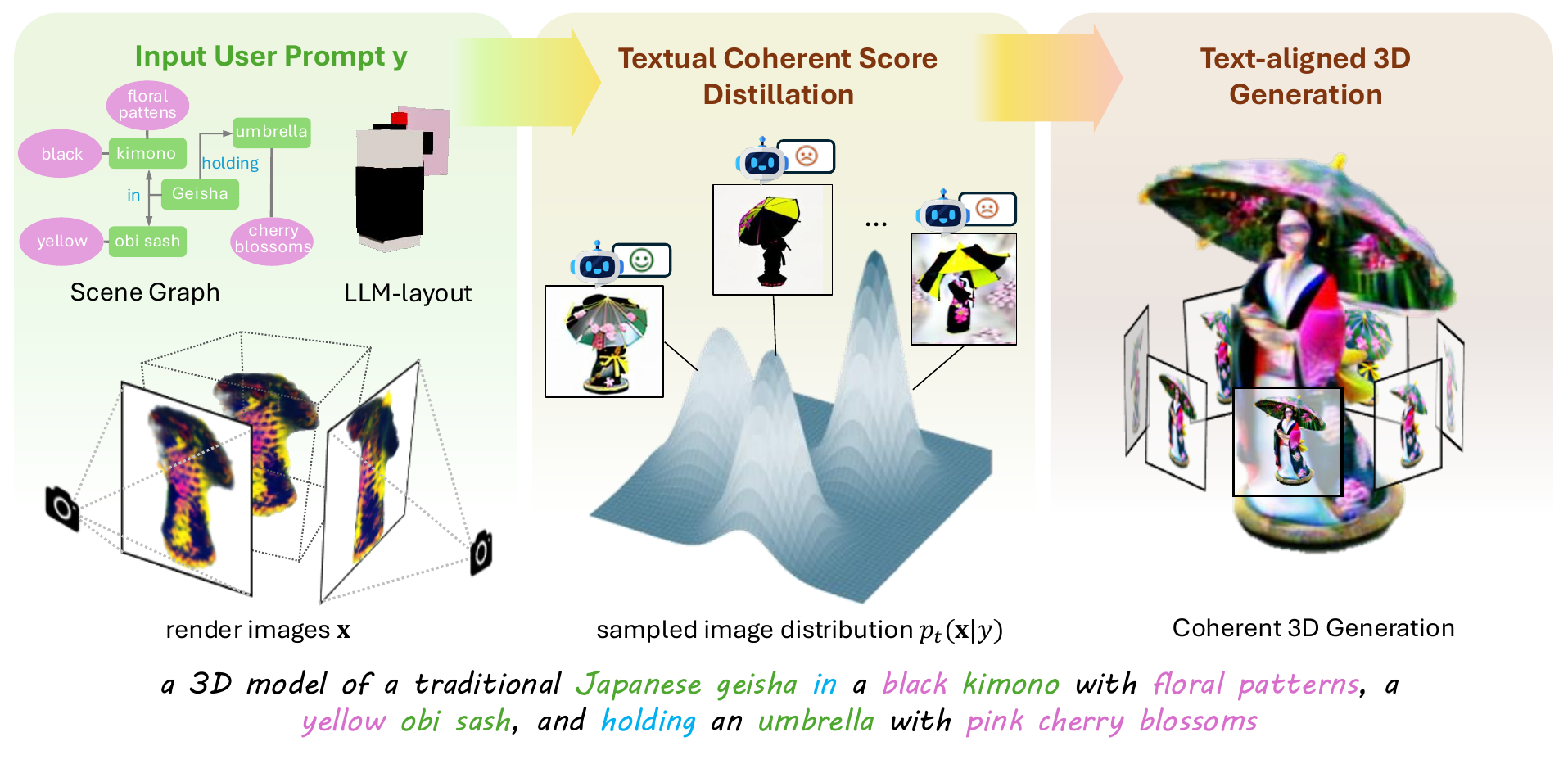}
  \vspace{-8mm}
  \caption{\textbf{Textual Coherent 3D Generation with CoherenDream.} By integrating multimodal LLM feedback into SDS optimization, our Textual Coherent Score Distillation corrects view-bias drift and yields faithful, text-aligned 3D content.}
  \label{fig:teaser}
\end{figure*}

\vspace{-6mm}
\input{sec/0_abstract} 
\begin{links}
    \link{Project Page}{https://chanyn.github.io/coherendream}
\end{links}

\input{sec/1_intro}
\input{sec/2_related}
\input{sec/3_preliminary}
\input{sec/4_method}
\input{sec/5_exp}

\input{sec/6_conclude}


\bibliography{aaai2026}

\input{supp}

\end{document}

%% file: sec/0_abstract.tex
\begin{abstract}
Score Distillation Sampling (SDS) has achieved remarkable success in text-to-3D content generation. However, SDS-based methods struggle to maintain semantic fidelity for user prompts, particularly when involving multiple objects with intricate interactions. 
While existing approaches often address 3D consistency through multiview diffusion model fine-tuning on 3D datasets, this strategy inadvertently exacerbates text-3D alignment degradation. 
The limitation stems from SDS's inherent accumulation of view-independent biases during optimization, which progressively diverges from the ideal text alignment direction.
To alleviate this limitation, we propose a novel SDS objective, dubbed as Textual Coherent Score Distillation (TCSD), which integrates alignment feedback from multimodal large language models (MLLMs). Our TCSD leverages cross-modal understanding capabilities of MLLMs to assess and guide the text-3D correspondence during the optimization. We further develop 3DLLaVA-CRITIC - a fine-tuned MLLM specialized for evaluating multiview text alignment in 3D generations. Additionally, we introduce an LLM-layout initialization that significantly accelerates optimization convergence through semantic-aware spatial configuration.
Our framework, CoherenDream, achieves consistent improvement across multiple metrics on TIFA subset.As the first study to incorporate MLLMs into SDS optimization, we also conduct extensive ablation studies to explore optimal MLLM adaptations for 3D generation tasks.

\end{abstract}

%% file: sec/1_intro.tex

\section{Introduction}
\label{sec:intro}
3D content generation is essential for diverse applications, including gaming, virtual reality, and robotics simulation. Recently, significant progress has been made in text-to-3D generation through  Score Distillation Sampling (SDS)~\cite{dreamfusion22,wang2023prolificdreamer,magic3d22,chen2023fantasia3d,shi2023mvdream,metzer2023latent}. SDS-based methods enable high-quality and diverse 3D generation based on user-provided text inputs, effectively distilling image distributions from 2D diffusion models~\cite{sd2022} into parameterized 3D representations.

Despite these advances, significant challenges persist in generating 3D content that adheres to user prompts, especially those involving multiple objects and intricate interactions. 
The limitation stems from SDS’s per-view distillation, which lacks global consistency constraints. Independent per-view optimization accumulates biases and progressively deviates from the intended text-to-3D alignment.
Furthermore, while recent variants of SDS~\cite{shi2023mvdream,liu2023syncdreamer,li2023sweetdreamer}  adopt fine-tuned diffusion models on specific 3D datasets to enhance 3D consistency, they exacerbate the problem of textual inconsistency in 3D generations~\cite{jiang2025jointdreamer,li2023sweetdreamer}, leading to object omissions and physically implausible spatial relationships. 

How to develop coherent text-to-3D generation remains relatively unexplored. Recent research efforts, such as JointDreamer~\cite{jiang2025jointdreamer}, emphasize the importance of maintaining textual consistency within the original diffusion model compared to its variants~\cite{shi2023mvdream}. DreamView~\cite{yan2025dreamview} trains a view-specific text-to-diffusion model to enhance alignment with specific viewpoints. However, both approaches still face the challenge of bias accumulation during SDS optimization.
Another intuitive way to alleviate textual inconsistency in 3D generation is through compositional strategies, such as GALA3D~\cite{zhou2024gala3d} and GraphDreamer~\cite{gao2024graphdreamer}. Nevertheless, GALA3D~\cite{zhou2024gala3d} suffers from non-overlap decomposition, which leads to unnatural spatial relationships. Although GraphDreamer~\cite{gao2024graphdreamer} advances the modeling of object relationships through scene graphs, 
it remains limited by compositional optimization, resulting in implausible object combination. 
In contrast, our work aims to generate coherent results within a holistic 3D representation where object relationships and spatial arrangements maintain textual consistency across all viewpoints.

Recent advances in multimodal large language models (MLLMs) demonstrate powerful cross-modal understanding capabilities, leading to successful integrations in text-to-image generation
~\cite{sun2023dreamsync,hu2024ella,feng2024ranni,lian2023llm,Cho2023VPT2I}. While 2D counterparts are promising, directly adapting MLLMs to 3D generation is not feasible, owing to their limited grasp of 3D representation and spatial relationships.
We address these limitations by reformulating text-3D alignment as across-view question-answering tasks. Our key insight is that MLLMs' semantic reasoning capabilities can complement gradient updates from image distribution when properly contextualized within SDS optimization, as shown in Fig~\ref{fig:teaser}. To this end, we present \textbf{CoherenDream}, which firstly regards MLLMs as dynamic semantic assessor to ensure textual consistency in SDS optimization.
%
We encode the input text and the across-view image distribution from the diffusion model as text-format ground truth, measuring the loss against the text sequences predicted by the MLLM to serve as feedback. This introduces Textual Coherent Score Distillation (TCSD), which adopts MLLM feedback in the SDS optimization to steer the optimization toward a textual-consistent distribution.


Current MLLMs are primarily designed as language assistants, they lack the proficiency to critique 3D generation effectively. To bridge this gap, we further develop a 3DLLaVA-CRITIC based on LLaVA-OneVision 0.5b~\cite{li2024llava} to enhance the quality of the feedback. Specifically, we design a view-aware data collection pipeline to simulate the gradient updates during SDS optimization, ensuring the delivery of accurate feedback. Furthermore, we introduce LLM-layout initialization, which is integrated with TCSD to warm up the 3D representation, thereby enhancing the textual consistency for image distribution from diffusion models. Extensive experimental results demonstrate that CoherenDream not only generates text-coherent 3D content but also outperforms other text-to-3D generation methods in terms of quality and quantitative metrics.
   
In summary, our contributions are as follows:
\begin{itemize}
    \item We introduce a novel Textual Coherent Score Distillation (TCSD) for text-coherent 3D generation, guiding optimization with MLLM feedback.
    \item We propose a view-aware data collection pipeline and fine-tune the 3DLLaVA-CRITIC to provide accurate feedback between text and 3D representations.
    \item Our CoherenDream establishes a new benchmark in coherent text-to-3D generation, producing 3D content that faithfully reflects user inputs.
\end{itemize}

%% file: sec/2_related.tex
\section{Related Works}
\label{sec:relat}
\paragraph{SDS-based Text-to-3D Generation.}
The Score Distillation Sampling (SDS) algorithm has achieved surprising results in text-to-3D generation. It pioneers by~\cite{dreamfusion22}, utilizing 2D diffusion model priors~\cite{sd2022} to optimize 3D representations. Recent advancements have further refined this technique by enhancing 3D representations~\cite{magic3d22, chen2023fantasia3d, yi2023gaussiandreamer, tang2023dreamgaussian}, improving generation quality~\cite{huang2023dreamtime, wang2023prolificdreamer, zhu2023hifa, liang2024luciddreamer}, and ensuring 3D consistency~\cite{jiang2025jointdreamer, shi2023mvdream, li2023sweetdreamer, seo2023let, perpneg23}. Despite these impressive results, these methods still struggle with multi-object prompts~\cite{he2023t} and semantic interaction. These challenges often stem from the lack of global textual consistency and accumulation of view-independent biases. To address it, 
we introduce MLLM feedback into SDS, which dynamically revises the update direction of 3D representations, guiding the score distillation process toward text-3D alignment.



\paragraph{Compositional Text-to-3D Generation.}
An intuitive approach for text-aligned 3D generation is to decompose holistic representations into individual components for separate optimization before recombination. However, existing methods~\cite{lin2023towards, bai2023componerf} suffer from quality degradation due to the challenges in managing layout constraints during NeRF optimization and their dependence on potentially inaccurate predefined layouts. Recently, GALA3D~\cite{zhou2024gala3d} attempts to use 3D Gaussians representation while dynamically refine layout during generation. However, its non-overlapping constraint leads to unnatural spatial relationships and scale inconsistencies. GraphDreamer~\cite{gao2024graphdreamer} advances the field by incorporating automatically parsed scene graphs to guide object interactions. Nevertheless, it inherits the fundamental limitations of compositional approaches, struggling to maintain holistic coherence, particularly in scenes requiring intimate object interactions.
%
%
In this work, we warm-up a holistic 3D representation through LLM-layout for SDS optimization. The proposed TCSD further enhance optimization towards text-3D alignment. 
Our approach can generate holistic 3D content while facilitating more realistic interactions among various objects.

\paragraph{Multimodal Large Language Models.}
Recent advancements in large language models (LLMs)\cite{touvron2023llama, openai2023gpt, anil2023palm} have led to increased interest in Multimodal Large Language Models (MLLMs), which combine vision understanding capabilities with LLMs\cite{openai2023gpt, li2024llava}. Given their promising abilities in visual reasoning and understanding, MLLMs are being explored to enhance 3D generation, particularly in areas such as automatic evaluation~\cite{wu2024gpt, he2023t} and data preparation~\cite{sun2024bootstrap3d, fang2024make}.
However, existing approaches have not yet integrated MLLMs directly into the 3D generation process. In contrast, our work fine-tunes the LLaVA model~\cite{li2024llava} to provide semantic coherence criticism and view-checking based on multi-view images. We are the first to use MLLM guidance to assist directly in SDS optimization.


%% file: sec/3_preliminary.tex
\section{Preliminaries}
\paragraph{Score Distillation Sampling. }
Score Distillation Sampling (SDS) employs priors from pre-trained 2D diffusion models to facilitate the generation of 3D content, which is widely adopted in text-to-3D methods~\cite{dreamfusion22,magic3d22,chen2023fantasia3d,wang2023prolificdreamer,luo2023diff}.
Given a user input prompt $y$, parameterized 3D representation $\theta$, and pre-trained 2D diffusion model $\Phi(\textbf{x}_t|y)$ along with noise prediction network $\epsilon_{\Phi}(\textbf{x}_t, t, y)$, the objective of SDS is to optimize $\theta$ by minimizing the KL-divergence between the rendered image distribution $q_t^{\theta}(\textbf{x}_t|c)$ and sampled image distribution $p_t(\textbf{x}_t|y))$ from diffusion model:
\begin{equation}
    \mathcal{L}_{SDS}(\theta)=\min\limits_{\theta}D_{KL}(q_t^{\theta}(\textbf{x}_t|c) || p_t(\textbf{x}_t|y)).
\label{eq:KLSDS}
\end{equation}
where $\textbf{x}_t$ represents the noisy rendered image $\textbf{x} = g(\theta, c)$ at timestamp $t$, and $g$ is the differentiable rendering function.

Ignoring the UNet Jacobian~\cite{dreamfusion22}, the gradient computation of SDS loss is as follows:
\begin{equation}\small
\begin{split}\label{eq:sds}
\nabla_{\theta}\mathcal{L}_{SDS}(\theta) & \triangleq \mathrm{E}_{t,\textbf{x}}[w(t)\frac{\sigma_t}{\alpha_t}\nabla_{\theta}D_{KL}(q_t^{\theta}(\textbf{x}_t|c, y)||p_t(\textbf{x}_t|y))]\\
    & \triangleq \mathrm{E}_{t,\epsilon_{\Phi}}[w(t)(\epsilon_{\Phi}(\textbf{x}_t, t, y)-\epsilon)\frac{\partial g(\theta, c)}{\partial \theta}],
\end{split}
\end{equation}
$\alpha_t$ and $\sigma_t$ are hyperparameters of noise schedule, $w(t)$ is the time-dependent weighting function, and $\epsilon_{\Phi} := (1+s)\epsilon_{\Phi}(\textbf{x}_t, t, y) - s\epsilon_{\Phi}(\textbf{x}_t, t, \emptyset)$ is the modification of noise prediction with classifier-free guidance (CFG) as $s$. 


%% file: sec/4_method.tex
\begin{figure*}[t]
\hsize=\textwidth 
  \centering
  \includegraphics[width=0.98\textwidth]{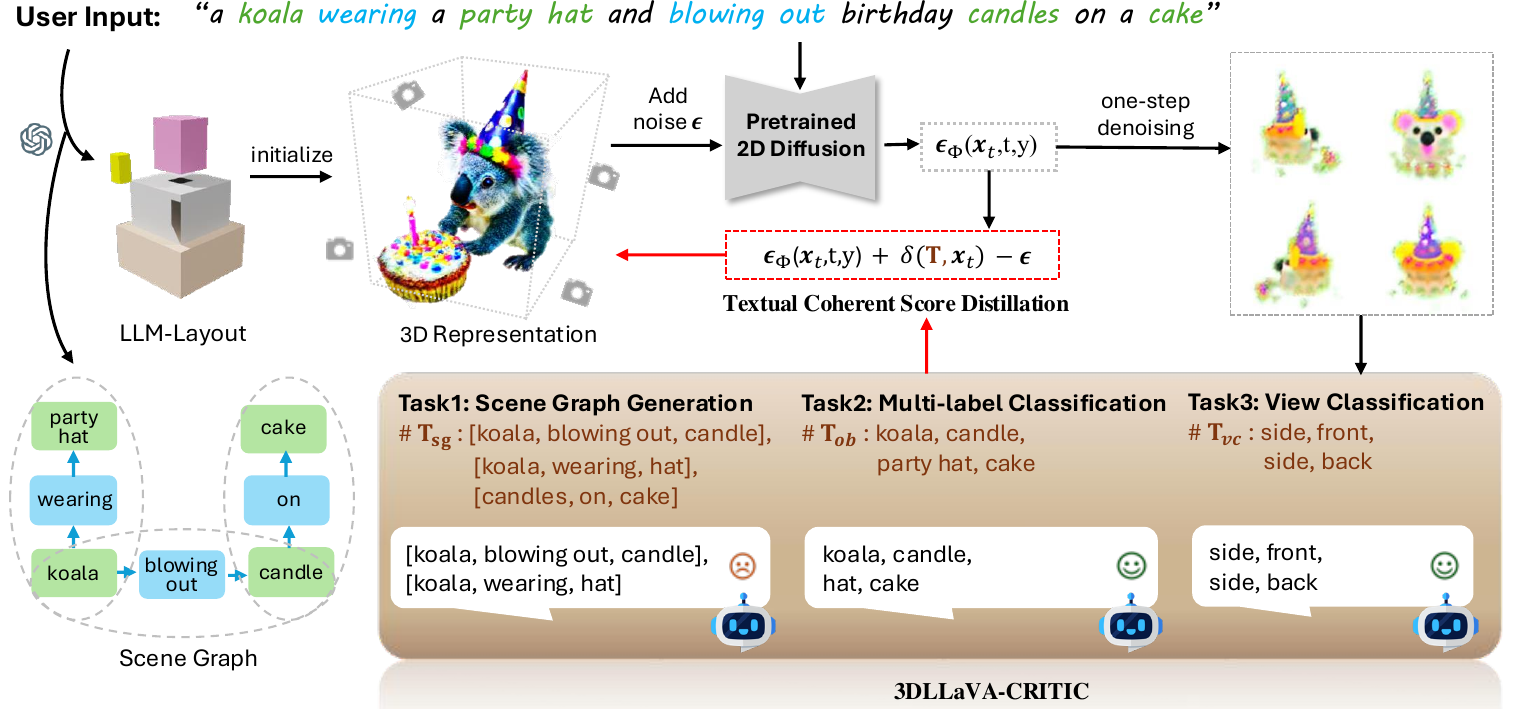}
   \vspace{-2mm}
   \caption{\textbf{Overview of CoherenDream framework.} CoherenDream involves LLM-Layout initialization, textual coherent score distillation and 3DLLaVA-CRITIC with three kinds of guidance tasks, producing text-3D aligned results from MLLM feedback.}
\label{fig:framework}
\vspace{-2mm}
\end{figure*}
\section{Method}
In this section, we introduce CoherenDream, a novel text-aligned 3D generation framework as depicted in~\cref{fig:framework}. 
We first demonstrate the derivation of Textual Coherent Score Distillation (TCSD), which introduce MLLM feedback for across-view image distribution. 
Then we introduce a fine-tuned MLLM, 3DLLaVA-CRITIC to better evaluate across-view alignment in 3D generation. Finally, we elaborate on the overall framework CoherenDream, where we integrate three guidance tasks with TCSD and novel LLM-layout initialization technique to further enhance textual consistency.

\subsection{Textual Coherent Score Distillation (TCSD)}
In accordance with~\cref{eq:KLSDS}, we observe significant limitations in textual understanding and reasoning within the original SDS framework: 
(1) $p_t(\textbf{x}_t|y)$ is constrained by diffusion models, which may diverge considerably from the actual user-prompt distributions~\cite{kirstain2023pick,sun2023dreamsync}; (2) $p_t(\textbf{x}_t|y)$ heavily relies on the rendered image $\textbf{x}$ of last updated 3D representation $\theta$, leading to bias accumulation. To address these limitations, we introduce informative feedback from MLLMs to $p_t(\textbf{x}_t|y)$ towards textual consistent distribution.

We first formulate the ideal textual consistent distribution $\hat{p}_t(\textbf{x}_t|y)$ as a joint distribution $\hat{p}_t(\textbf{x}_t|y)=p_t(\textbf{x}_t|y)p(y|\textbf{x}_t)$.
Based on the connection between diffusion models and score matching~\cite{song2019generative}, the score function of $p_t(\textbf{x}_t|y)$ can be derived as:
\begin{equation}\label{eq:score_fuction}
\nabla_{\textbf{x}_t}\text{log}\, p_t(\textbf{x}_t|y) = -\frac{1}{\sqrt{1-\alpha_t}}\epsilon_\Phi(\textbf{x},t,y)
\end{equation}
We substitute~\cref{eq:score_fuction} into score function for $\hat{p}_t(\textbf{x}_t|y)$:
\begin{equation}\small
\nabla_{\textbf{x}_t}\text{log}( p_t(\textbf{x}_t|y)p(y|\textbf{x}_t)) = -\frac{1}{\sqrt{1-\alpha_t}}\epsilon_\Phi(\textbf{x},t,y)+\nabla_{\textbf{x}_t}\text{log}\,p(y|\textbf{x}_t)
\end{equation}
Then, we derive the textual consistent noise prediction:
\begin{equation}
\begin{split}\small\label{eq:epsilon}
\hat{\epsilon}_\Phi(\textbf{x}_t, t, y)=\epsilon_\Phi(\textbf{x}_t, t, y)-\sqrt{1-\alpha_t}\nabla_{\textbf{x}_t}\text{log}\,p(y|\textbf{x}_t)\\
=\epsilon_\Phi(\textbf{x}_t, t, y)+\lambda\cdot\nabla_{\textbf{x}_t}\mathcal{L}_{ce}(\textbf{T},f_{cr}(\textbf{x}_t))
\end{split}
\end{equation}
where $\lambda=\sqrt{1-\alpha_t}$ controls the guidance strength, $\mathcal{L}_{ce}$ denotes autogressive cross-entropy loss,  $f_{cr}$ indicates MLLM as guidance function, and $\textbf{T}$ refers to predefined questions and answers related to $y$. However, deploying guidance on noisy input $\textbf{x}_t$ is not practical, we consider calculating feedback on one-step denoising data $\hat{\textbf{x}}_0$. Hence, the final feedback $\delta(\textbf{T}, \hat{\textbf{x}}_0)$ for textual consistency is represented as:
\begin{equation}
\begin{split}\small\label{eq:feedback}
\delta(\textbf{T},\hat{\textbf{x}}_0)& =\nabla_{\textbf{x}_t}\mathcal{L}_{ce}(\textbf{T},f_{cr}(\hat{\textbf{x}}_0))\\
& =\nabla_{\textbf{x}_t}\mathcal{L}_{ce}(\textbf{T},f_{cr}(\frac{\textbf{x}_t-\sqrt{1-\alpha_t}\epsilon_\Phi(\textbf{x}_t,t,y)}{\sqrt\alpha}))
\end{split}
\end{equation}

Thus, the  textual consistent gradient of Textual Coherent Score Distillation can be reformulated by~\cref{eq:sds} and ideal noise prediction in~\cref{eq:epsilon} as follows:
\begin{equation}
\begin{split}\small\label{eq:tcsd}
\nabla_{\theta}\mathcal{L}_{TCSD}(\theta)\triangleq \mathrm{E}_{t,\epsilon_{\Phi}}[w(t)(\hat\epsilon_{\Phi}(\textbf{x}_t, t, y)-\epsilon)\frac{\partial g(\theta, c)}{\partial \theta}],\\
\triangleq \mathrm{E}_{t,\epsilon_{\Phi}}[w(t)(\epsilon_{\Phi}(\textbf{x}_t, t, y) + \lambda\cdot\delta(\textbf{T},\hat{\textbf{x}}_0) -\epsilon)\frac{\partial g(\theta, c)}{\partial \theta}],
\end{split}
\end{equation}

\subsection{3DLLaVA-CRITIC}
\label{sec:3dllava}
Original MLLMs are primarily designed as language assistants and cannot assess 3D generation adequately. To bridge this gap, we decompose the evaluation of textual consistency in 3D generation into three question-answering tasks: scene graph generation, multi-label image classification, and view classification. Building on this foundation, we introduce 3DLLaVA-CRITIC, following the architecture of 
LLaVA-OV~\cite{li2024llava} and fine-tuned using instruct-tuning pairs generated by GPT-4o~\cite{openai2023gpt}. Detailed prompts can be found in the the Appendix.

\paragraph{Task Definition }
\begin{itemize}
\item \textbf{Scene graph generation.} We establish this task to enhance global semantic understanding in across-view image distribution. Unlike caption generation, which often includes excessive and irrelevant descriptive details, the scene graph format requires the model to focus on object interactions. Specifically, given across-view observation, the 3DLLaVA-CRITIC produces a set of scene graphs represented as triplets. The form of scene graph triplet is \texttt{[subject, relation, object]} or \texttt{[object, is, attribute]}. To avoid ambiguity, we define four types of relation: Actions, Spatial Relations, Descriptive Verbs, and Non-specific Connections (e.g., ``and").
\item \textbf{Multi-label classification.} We observe that missing objects are a common phenomenon in textual inconsistency. Therefore, multi-label classification task asks 3DLLaVA-CRITIC to focus on primary objects presented in the user's input. Specifically, 3DLLaVA-CRITIC need to extract objects in given observation.
\item \textbf{View classification.} Given input images, the 3DLLaVA-CRITIC determines the input camera position from the options: \texttt{side, front, back, overhead.} By comparing this classification with the sampled camera positions, the model guides the SDS to generate results that are consistent with the correct viewpoints.
\end{itemize}

\paragraph{View-aware Data Collection Pipeline}
Unlike natural images, the sampled image distribution in SDS is conditioned on noisy rendered images from randomly sampled camera poses. To achieve better assessment during SDS optimization, we introduce a view-aware data collection pipeline, as illustrated in~\cref{fig:data}. Detailed prompts for each step in the pipeline are available in the Appendix.

\noindent\textbf{Layout condition generation.} Inspired by LayoutGPT~\cite{feng2024layoutgpt}, we utilize GPT-4~\cite{openai2023gpt} to generate 200 diverse text prompts based on examples in the DreamFusion~\cite{dreamfusion22} library. Layout is defined as 3D bounding boxes with box center coordinates and dimension: \texttt{(x, y, z, h, w, l)}. These layouts are rendered in Blender, and we randomly sample camera poses to produce 32 rendering images per layout.

\noindent\textbf{View-aware image generation.} Based on the previously created prompts and layout condition images, we enhance the original prompts by integrating view prompts (e.g., \texttt{side, back, front, overhead}). Utilizing DeepFloyd-IF~\cite{deepfloyd} and MVDream~\cite{shi2023mvdream}, we generate view-aware images by varying random seeds, noise strengths, and denoising steps, in line with the optimization process of CoherenDream. Under each layout condition, we randomly select 4 camera views to create a 2$\times$2 grid image, resulting in a total of 36 grid images.

\noindent\textbf{GPT-4o Annotation.} 
After the last step, we curate 307,409 grid images and then prompt GPT-4o~\cite{openai2023gpt} to generate scene graphs, view prompts, and perform object extraction. These high-quality grid image-text pairs serve as instruction tuning data for proposed 3DLLaVA-CRITIC.


\begin{figure}[t]
  \centering
\includegraphics[width=0.48\textwidth]{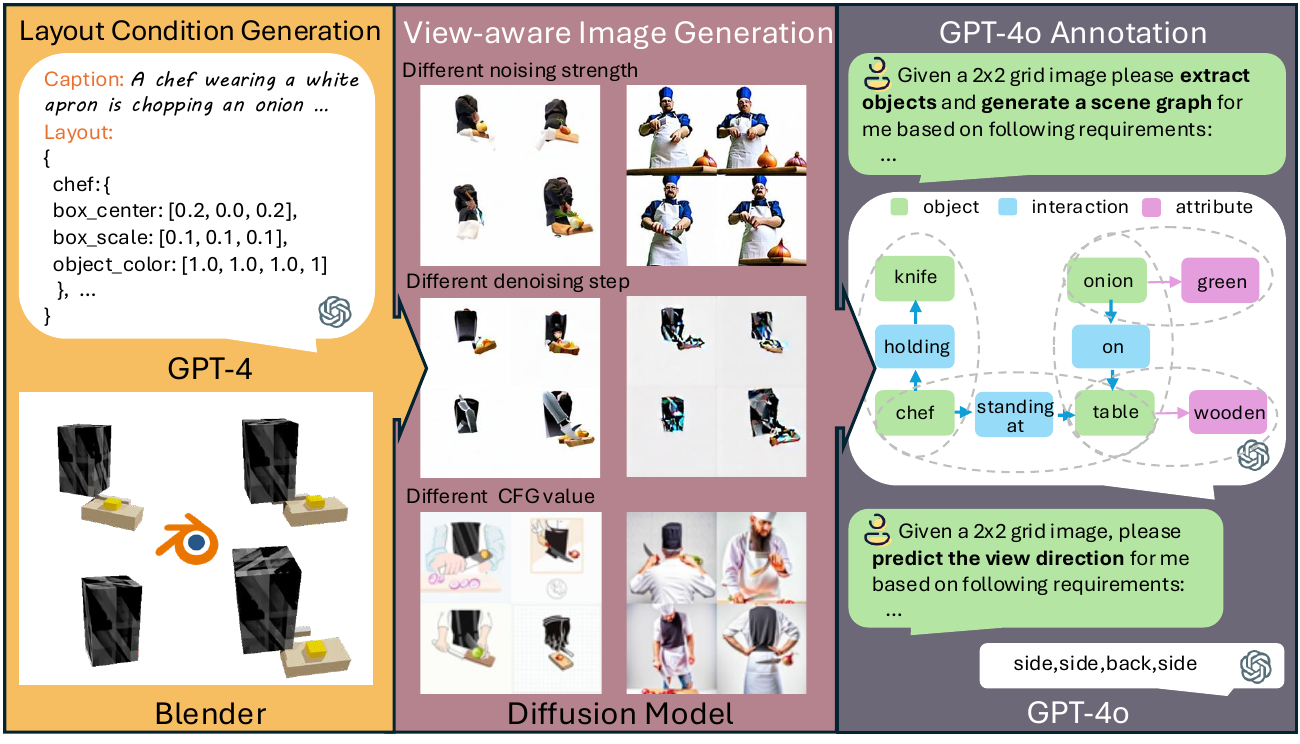}
   \caption{\textbf{View-aware data collection pipeline for 3DLLaVA-CRITIC} that consist of (1) using LLM to generate diverse coarse text prompt and corresponding layout and rendering from random viewpoints in Blender; (2) random sampling images from T2I diffusion model conditioned on layout image and randomly assembling into a 2$\times$2 grid image; (3) employing GPT-4o to extract semantic annotation, including scene graph and view direction.
   }
\label{fig:data}
\vspace{-2mm}
\end{figure}

\subsection{Framework of CoherenDream}\label{sec:framework}
Building upon TCSD optimization, we present the CoherenDream framework that is based on NeRF~\cite{mildenhall2021nerf} as 3D representation, utilizing Instant-NGP~\cite{muller2022instant} with a volume renderer. To ensure coherence with user input, we incorporate three guidance tasks 
with~\cref{eq:tcsd}, which direct sampled image distribution toward the textual consistent distribution. During the optimization process, we employ well-established techniques such as time-annealing~\cite{huang2023dreamtime} and resolution scaling-up~\cite{wang2023prolificdreamer}. In addition, we introduce a novel technique called LLM-layout initialization, which significantly enhances the quality of the generated 3D content.

\paragraph{Guidance Tasks. } We equipe TCSD optimization with three feedbacks from 3DLLaVA-CRITIC: in scene graph generation, multi-label classification and view classification. Detailed task descriptions can be found before.
For the scene graph generation task, $\textbf{T}_{sg}$ can be extracted by humans or GPT according to user's input. And answers $\textbf{T}_{ob}$ of multi-label classification can be directly obtained from $\textbf{T}_{sg}$. $\textbf{T}_{view}$ can be decided given the sampled camera pose. 
Therefore, the overall feedback of CoherenDream is demonstrated as: $\delta_{\text{CoherenDream}}=\lambda_{sg}\delta(\textbf{T}_{sg},\hat{\textbf{x}}_0)+\lambda_{ob}\delta(\textbf{T}_{ob},\hat{\textbf{x}}_0)+\lambda_{view}\delta(\textbf{T}_{view},\hat{\textbf{x}}_0)$.


\begin{figure*}[t]
\hsize=\textwidth 
  \centering
  \includegraphics[width=0.98\textwidth]{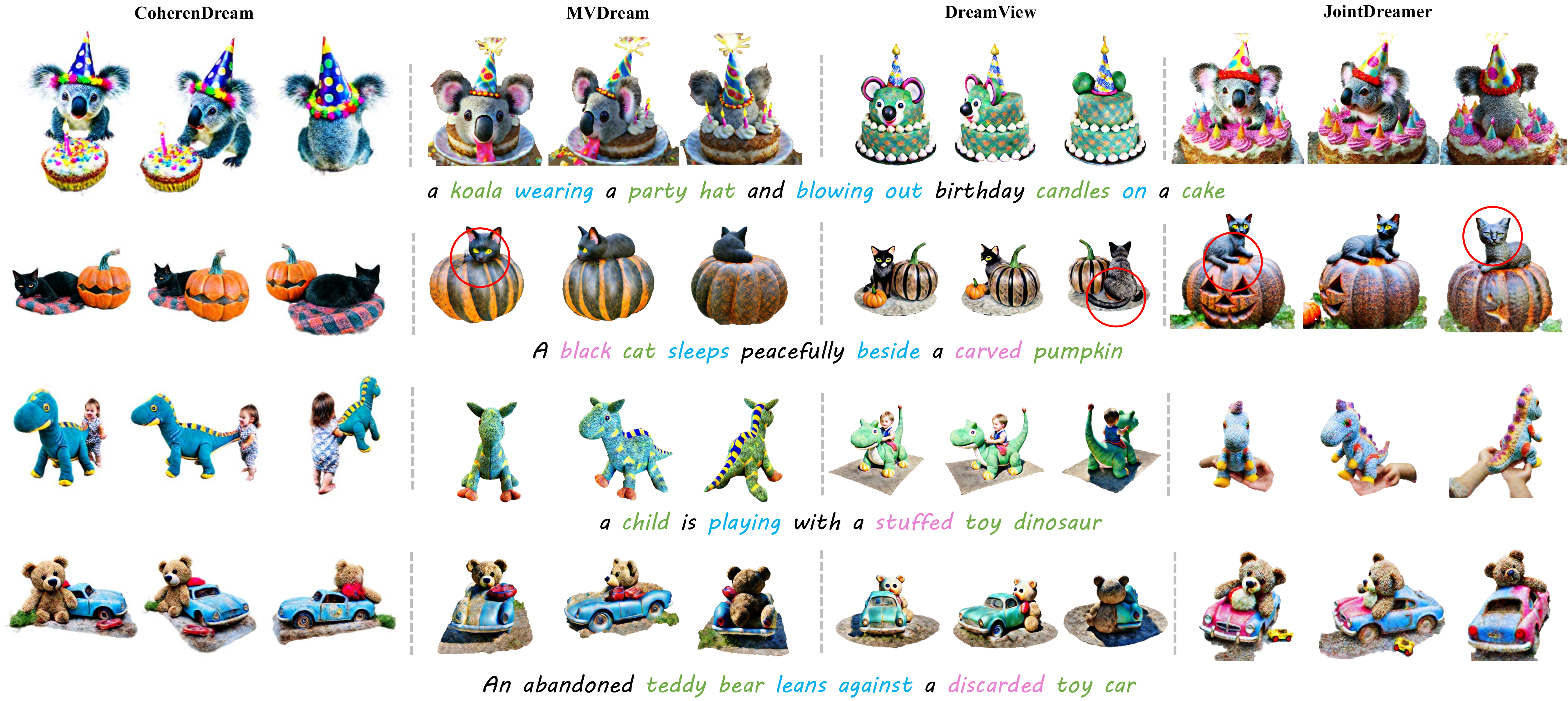}
   \vspace{-2mm}
   \caption{\textbf{Qualitative comparison with representative methods.} The results indicate that existing text-to-3D generation methods do not produce textual consistent results, involving objects omissions or unnatural interactions (highlighted in red). Conversely, our CoherenDream generates more textually faithful results that benefited from effective MLLM feedback.}
\label{fig:compared_methods}
\end{figure*}

\paragraph{LLM-layout Initialization. }
As discussed in Textual Coherent Score Distillation, the rendered images $\textbf{x}$ of 3D representations significantly influence the sampled image distribution from the diffusion model. The usage of a unit sphere for initialization provides less information condition, leading to the traption of local minimum during the score distillation optimization. It also leads to textual inconsistency. To address this issue, we introduce an LLM-generated layout to warm up CoherenDream generation, which we refer to as LLM-Layout. As illustrated in the ``Layout Condition Generation" part of~\cref{fig:data}, LLM-Layout consists of a set of cubes in normalized space. The collection of LLM-Layouts mirrors the layout condition generation process detailed in 3DLLaVA-CRITIC.
Specifically, we fit the density network of NeRF to align with the occupancy defined by the LLM-Layout. Since the LLM-Layout only provides fundamental placement information, we do not wish to constrain the 3D representation's flexibility regarding shape. To balance this, we implement a decay in importance near the surface, allowing for more freedom in shape representation. The surface weight decay binary cross-entropy loss function is:
\begin{equation*}
    \mathcal{L}_{\text{LLM-Layout}}=\mathcal{L}_{bce}(\text{occ}_{\theta}(p), \text{occ}_{\text{LLM-Layout}}(p))\dot(1-e^{\frac{d^2}{2\sigma}})
\end{equation*}
where $\text{occ}_\theta$ indicates occupancy prediction from density network of NeRF $\theta$, $p$ is the points set sampled from camera view, $d$ denotes the distance of $p$ from the surface, and $\sigma$ is a hyperparameter
that controls the strength of the constraint. During the warming-up phase, which consists of $N=600$ steps, the noising strength is restricted to the range  $[0.4, 0.7]$ to encourage the sampled distribution to adhere more closely to the initialized rendered image. Additionally, the importance of $\mathcal{L}_{\text{LLM-Layout}}$ decays over the warming-up steps. This approach enables us to achieve a better initialization for the SCSD optimization without severely restricting the representational capacity of NeRF.

%% file: sec/5_exp.tex
\section{Experiments}
\label{sec:exp}
We present the text-to-3D generation results of CoherenDream with qualitative and quantitative evaluations, illustrating state-of-the-art performance. We also make an ablation analysis of the proposed TCSD and 3DLLaVA-CRITIC. More details and experiments can be found in the Appendix.

\subsection{Textual Consistent 3D Generation}
\paragraph{Qualitative Comparison. } Figure~\ref{fig:compared_methods} shows the comparison with several representative baselines, results produced by official codes.
\textit{(\romannumeral1) MVDream~\cite{shi2023mvdream}:} While MVDream finetunes multi-view diffusion model on 3D dataset demonstrating heavy textual inconsistency with multi-object prompts, our CoherenDream triggers generalization of multi-view diffusion model guiding by MLLMs feedback. 
\textit{(\romannumeral2) DreamView~\cite{yan2025dreamview}:} DreamView finetunes on 3D dataset leading to unrealistic texture and implausible interaction (e.g., the mixture of ``bear" and ``toy car").
Additionally, it need view-specific prompts, which cost human labor.
\textit{(\romannumeral3) JointDreamer~\cite{jiang2025jointdreamer}:} JointDreamer maintains the generalization of original 2D diffusion, but it cannot break through the inherent limitations of diffusion model. It demonstrates object omissions (``child" in the third line) and insufficient semantic interaction understanding (targets for ``beside" but shows ``on" in the second row). In comparison, our Conherent can produce more text-faithful results.

\input{tab/compare}
\input{tab/3dlavacritic}

\noindent\textbf{Quantitative Comparison. } We perform quantitative evaluations on a curated 45-prompt subset of TIFA v1.0~\cite{hu2023tifa}. 
We adopt a multi-metric evaluation protocol to evaluate textual consistency:
TIFA score~\cite{hu2023tifa}, VQAScore~\cite{lin2024evaluating} and CLIP score~\cite{hessel2021clipscore}.To ensure comprehensive viewpoint coverage, we uniformly sample 10 azimuth angles around each 3D asset. For VQAScore, we report the maximum score across all viewpoints; For TIFA Score, we compute the intersection of correct answers across all viewpoints. For Clip Score, we adopt the CLIP ViT-B/32 as the feature extractor and calculate the average score across views. Quantitative results in Table~\ref{tab:quantitative1} demonstrate the consistent superiority of CoherenDream in text-aligned 3D generation across all metrics. Specifically, CoherenDream achieves an improvement of the TIFA score by 4.0\% over MVDream, demonstrating its superior corresponding to textual description benefiting from effective feedback from MLLM. 

\subsection{Ablation Study}

\begin{figure}[t]
\vspace{-4mm}
  \centering
  \includegraphics[width=0.48\textwidth]{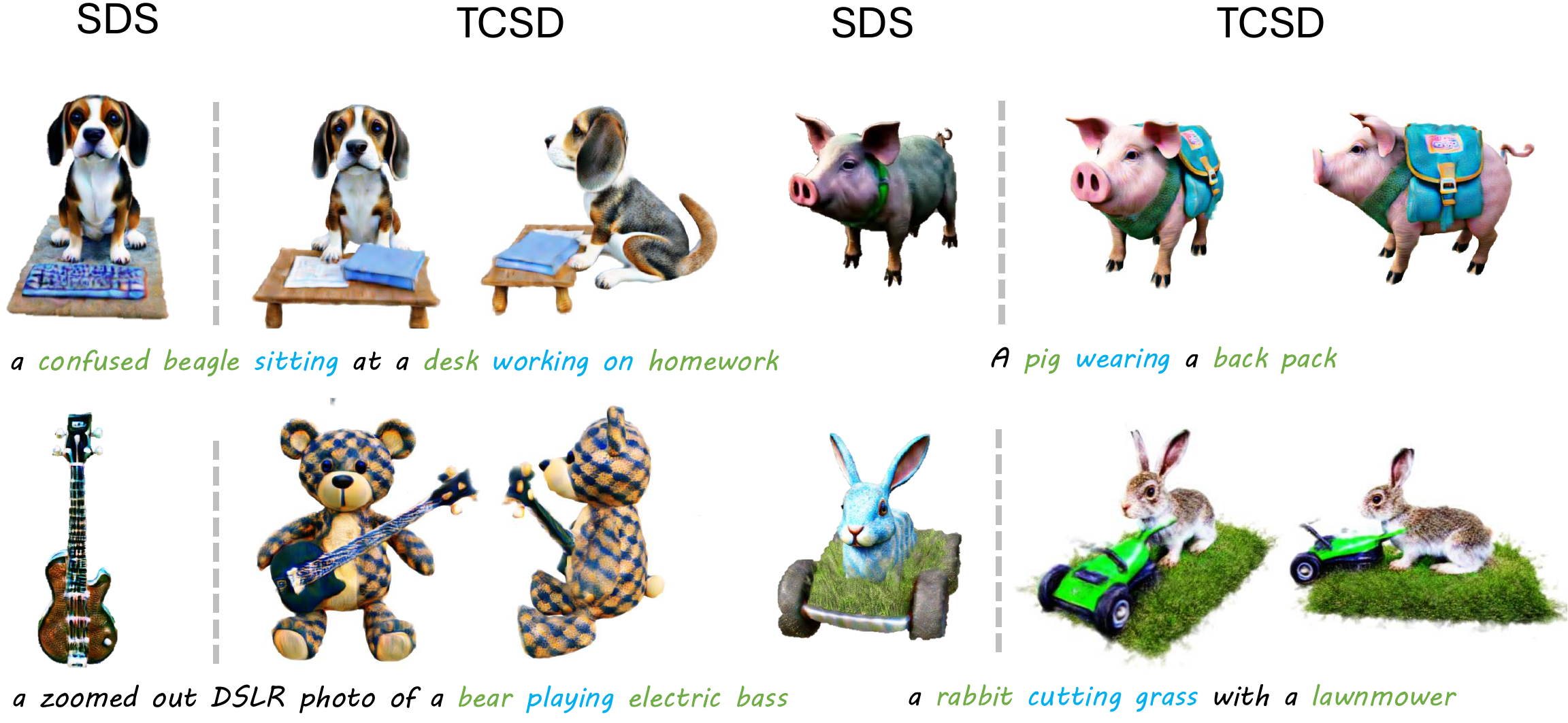}
   \caption{\textbf{Comparing TCSD with Original SDS.} Original SDS exhibits bias accumulation issues resulting in object omissions, while TCSD leverages a dynamic MLLM assessor to produce coherent results in a holistic 3D space.}
\label{fig:compare_baseline}
\vspace{-3mm}
\end{figure}

\noindent\textbf{Ablation on TCSD.} To verify the effectiveness of text-to-3D alignment achieved by TCSD, we distill from the multi-view diffusion model used in MVDream~\cite{shi2023mvdream}, which is known to overfit on 3D datasets. And we only utilize the original LLaVA-ov-0.5b with scene graph generation task as guidance. The primary distinction between the original SDS and TCSD, is the introduction of an optimization direction $\delta$, which aims for ideal text alignment distribution defined in~\cref{eq:feedback}. Original SDS lacks the understanding and reasoning capabilities necessary for proper alignment, leading to significant misalignment issues. The incorporation of $\delta$ serves to mitigate these issues. The results presented in~\cref{fig:compare_baseline} validate our claims regarding the effectiveness of this approach.

\begin{figure*}[t]
\hsize=\textwidth 
  \centering
  \includegraphics[width=0.98\textwidth]{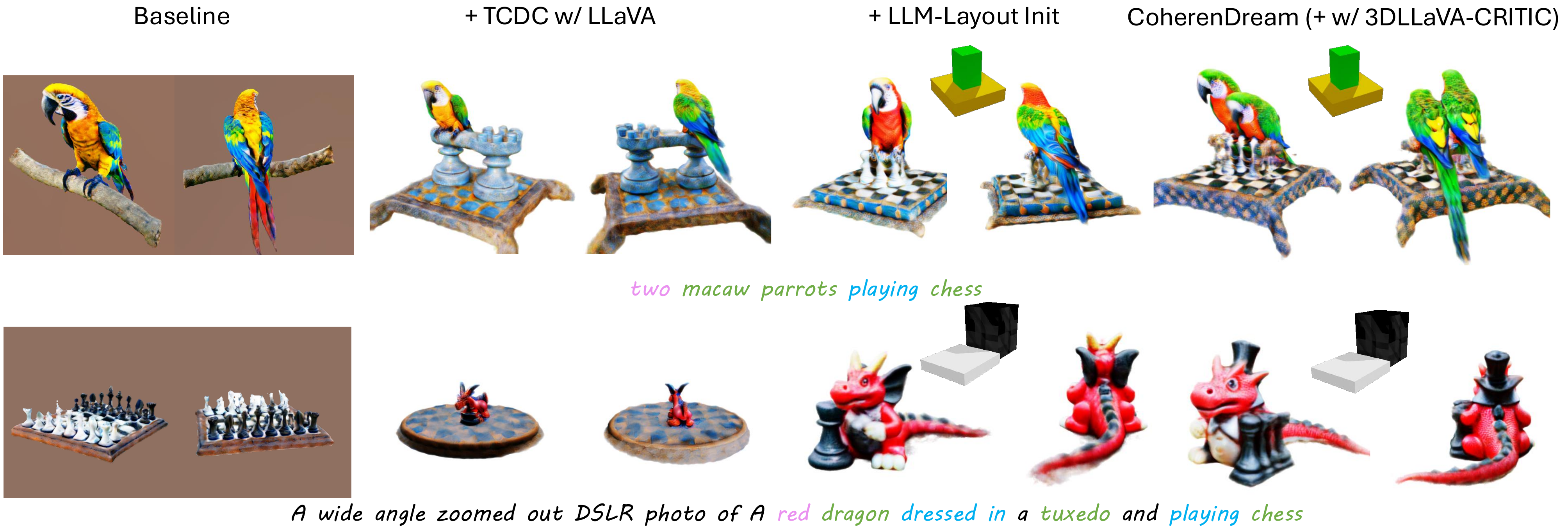}
   \vspace{-2mm}
   \caption{\textbf{Incremental ablations on techniques in CoherenDream framework}, which enhances text alignment.}
\label{fig:ablation}
\end{figure*}

\begin{figure*}[t]
\hsize=\textwidth 
  \centering
  \includegraphics[width=0.98\textwidth]{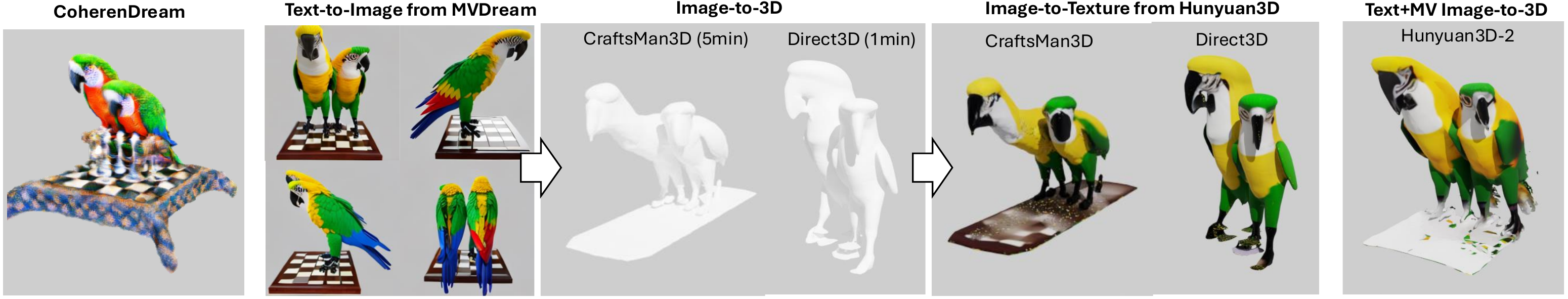}
   \caption{\textbf{Compare with native 3D model in text-to-image-to-3D pipeline.} }
\label{fig:compare_with_I23D}
\vspace{-4mm}
\end{figure*}

\paragraph{Ablation on LLM-Layout Initialization and  3DLLaVA-CRITIC. }We conduct incremental ablations on techniques in CoherenDream, focusing on LLM-Layout Initialization and the fine-tuned 3DLLaVA-CRITIC. As shown in~\cref{fig:ablation}, LLM-Layout Initialization enhances the warm-up process for the 3D representation, effectively activating semantic information within the diffusion model and preventing the optimization process from collapsing into local minima. Compared to the scenario without LLM-Layout, this technique helps the model focus on the primary objects and establishes reasonable relative scale relationships. 
Moreover, the proposed LLM-Layout Initialization strategy does not constrain the creativity of 3D presentations. 

However, the domain gap between real images and diffusion‐sampled views causes off‐the‐shelf MLLMs to miss fine attributes and objects (e.g., ignore ``a parrot is two” or ``tuxedo” as shown in Fig.~\ref{fig:ablation}. To address this, we fine‐tune 3DLLaVA-CRITIC, which delivers more accurate text‐to‐3D feedback and yields significantly more faithful 3D outputs. We also quantitatively compare 3DLLaVA-CRITIC against LLaVA-OV-0.5B in Table \ref{tab:supp-computation}. In the absence of a ground-truth dataset, we construct 30 prompts for $\textbf{T}_{sg}$ and $\textbf{T}_{ob}$, and randomly choose 10 objects across 10 camera poses for $\textbf{T}_{view}$. For each validation sample, GPT-4o is given both models’ responses alongside the reference answer and tasked with scoring each as correct (1) or incorrect (0). The resulting average accuracies confirm the consistent gains achieved by our fine-tuning.

\paragraph{Compare with native 3D models. } 
Recently, native 3D generation models have attracted attention due to their rapid inference. However, constrained by the limited size and diversity of existing 3D datasets, these native models underperform optimization‐based approaches on complex or lengthy textual prompts. Fig.~\ref{fig:compare_with_I23D} presents a qualitative comparison between our method and several native pipelines, including Craftsman3D \cite{li2024craftsman3d}, Direct3D \cite{wu2024direct3d}, and Hunyuan3D-2 \cite{zhao2025hunyuan3d}.
Most native models do not generate 3D geometry directly from text; instead, they employ a multistage text-image-textured-mesh pipeline. As a result, the final mesh quality depends heavily on the intermediate image and often omits fine details. Furthermore, errors introduced during text‐to‐image synthesis propagate through each stage, leading to significant misalignment between the user’s prompt and the native 3D output. These pipelines also tend to produce less realistic textures than SDS-based methods. Finally, although native models offer fast image‐to‐3D conversion, applying textures to an initially untextured mesh remains a computationally expensive step. In contrast, our CoherenDream maintains higher fidelity to textual prompts and produces more detailed, semantically consistent 3D content.

%% file: tab/compare.tex
\begin{table}[t]
	\begin{center}
\resizebox{0.48\textwidth}{!}{
    \begin{centering}
    \begin{tabular}{c|ccc}
\toprule

Method & TIFA Score $\uparrow $& VQAScore $\uparrow $ & CLIP Score $\uparrow$\\
\midrule 
MVDream~\cite{shi2023mvdream} & 77.4 & 0.73 & 30.5 \\
DreamView~\cite{yan2025dreamview} & 77.9 & 0.71 & 31.1 \\
JointDreamer~\cite{jiang2025jointdreamer}  & 73.7 & 0.71& 30.8\\
CoherenDream  & \textbf{81.4} &  \textbf{0.79}& \textbf{31.4}\\
\bottomrule
\end{tabular}
\end{centering}
}

\end{center}
\vspace{-4mm}
\caption{\textbf{Comparison on TIFA V1.0 subset.}}
\label{tab:quantitative1}
\vspace{-4mm}
\end{table}

%% file: tab/3dlavacritic.tex
\begin{table}[t]
\begin{center}
\resizebox{0.48\textwidth}{!}{
    \begin{centering}
\begin{tabular}{c|ccc}
\toprule
Method & $Acc_{\textbf{T}_{sg}}\uparrow $ & $\textbf{T}_{ob}$$\uparrow $ & $\textbf{T}_{view}$$\uparrow $ \\
\midrule
LlaVA-OV-0.5B~\cite{li2024llava} &  0.21  &  0.34  &  0.72 \\
3DLlava-CRITIC &  0.76  &  0.89  &  0.87 \\
\bottomrule         
\end{tabular}
\end{centering}
}

\end{center}
\vspace{-4mm}
\caption{\textbf{Quantitative evaluation for 3DLlava-CRITIC.}}
\label{tab:supp-computation}
\vspace{-4mm}
\end{table}

%% file: sec/6_conclude.tex
\section{Conclusion}
In this paper, we introduce CoherenDream, a novel text-to-3D framework that leverages a powerful Multimodal Large Language Model (MLLM) to generate 3D results that are faithful to the user's inputs. We demonstrate the framework's effectiveness in optimizing score distillation through semantic feedback derived from the MLLM. Additionally, we incorporate LLM-Layout Initialization to enhance the warm-up process of the 3D representation.
Our experiments show that CoherenDream achieves state-of-the-art performance in generating multi-object 3D content, excelling in both visual appearance and alignment with input text.

\textbf{Limitation.} CoherenDream excels at holistic 3D generation and capturing intricate object interactions, but may underperform compositional methods in precise object arrangement and partial attribute control. But our methods can be integrated with compositional pipelines to harness complementary strengths. 

%% file: supp.tex
\clearpage
\setcounter{page}{1}

This supplementary material consists of three parts, including technical details of the experimental setup, prompt design and additional experimental results.

\section{Implementation Details}
\label{sec:supp-detail}
In this section, we present comprehensive details regarding the training processes and data utilized in 3DLLaVA-CRITIC, as well as the implementation specifics related to the text-to-3D generation framework of CoherenDream and the baseline methods employed.

\subsection{Details of 3DLLaVA-CRITIC}
\begin{figure}[h]
  \centering
  \includegraphics[width=0.5\textwidth]{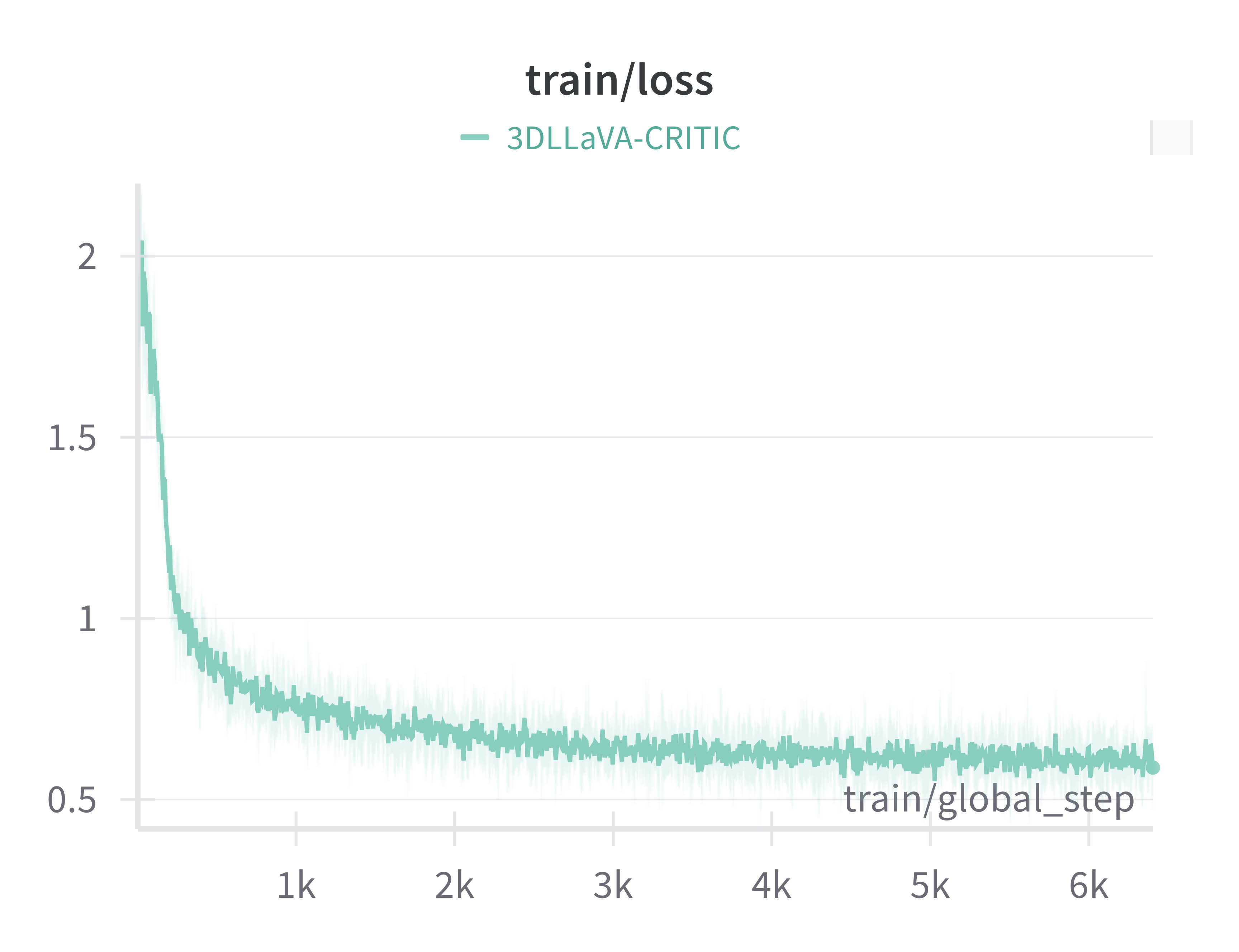}
  \vspace{-6mm}
   \caption{\textbf{Training loss of 3DLLaVA-CRITIC.} }
\label{fig:supp-llavaloss}
\vspace{-4mm}
\end{figure}

Based on the collected view-aware dataset, we finetune the LLaVA-OneVision 0.5B~\cite{li2024llava} model. We use 8 RTX 4080 to train the model. The batch size is 32 (4 grid images across 8 GPUs), and the gradient is accumulated from 2 batches, forming a total batch size of 64. We use the AdamW [29] optimizer, the learning rate for Lora~\cite{hu2021lora} and multimodal MLP adapter is set to 2e-5, and the learning rate for the vision tower is set to 2e-6. The image size is 384$\times$384. The fine-tuning takes 3.5 hours and the loss curve can be found in~\cref{fig:supp-llavaloss}.

\paragraph{View-aware Image Generation Details.}
We construct a view-aware image dataset to bridge the gap between natural images and the conditional image distributions sampled from diffusion models during SDS optimization. This dataset is designed based on a configuration similar to that used in SDS optimization, utilizing samples from MVDream~\cite{shi2023mvdream} and DeepFloy-IF~\cite{deepfloyd}.
For each layout image, we sample four random seeds and apply three different noise strengths, drawn from the intervals (0.02, 0.3), (0.3, 0.6), and (0.5, 0.8). Additionally, we implement three denoising steps with values of 50, 20, and 5. The classifier-free guidance (CFG) values are randomly selected from the ranges [30, 75] for MVDream and [8, 20] for DeepFloy-IF. 
Furthermore, the overall text descriptions integrate view-specific information (e.g., side, back, front, overhead) based on the camera matrix saved during the layout generation phase. This integration provides a more comprehensive context for the generated images. Notably, the camera matrix is also utilized as a condition during the sampling process for MVDream.

\begin{figure*}[t]
\hsize=\textwidth 
  \centering
  \includegraphics[width=0.98\textwidth]{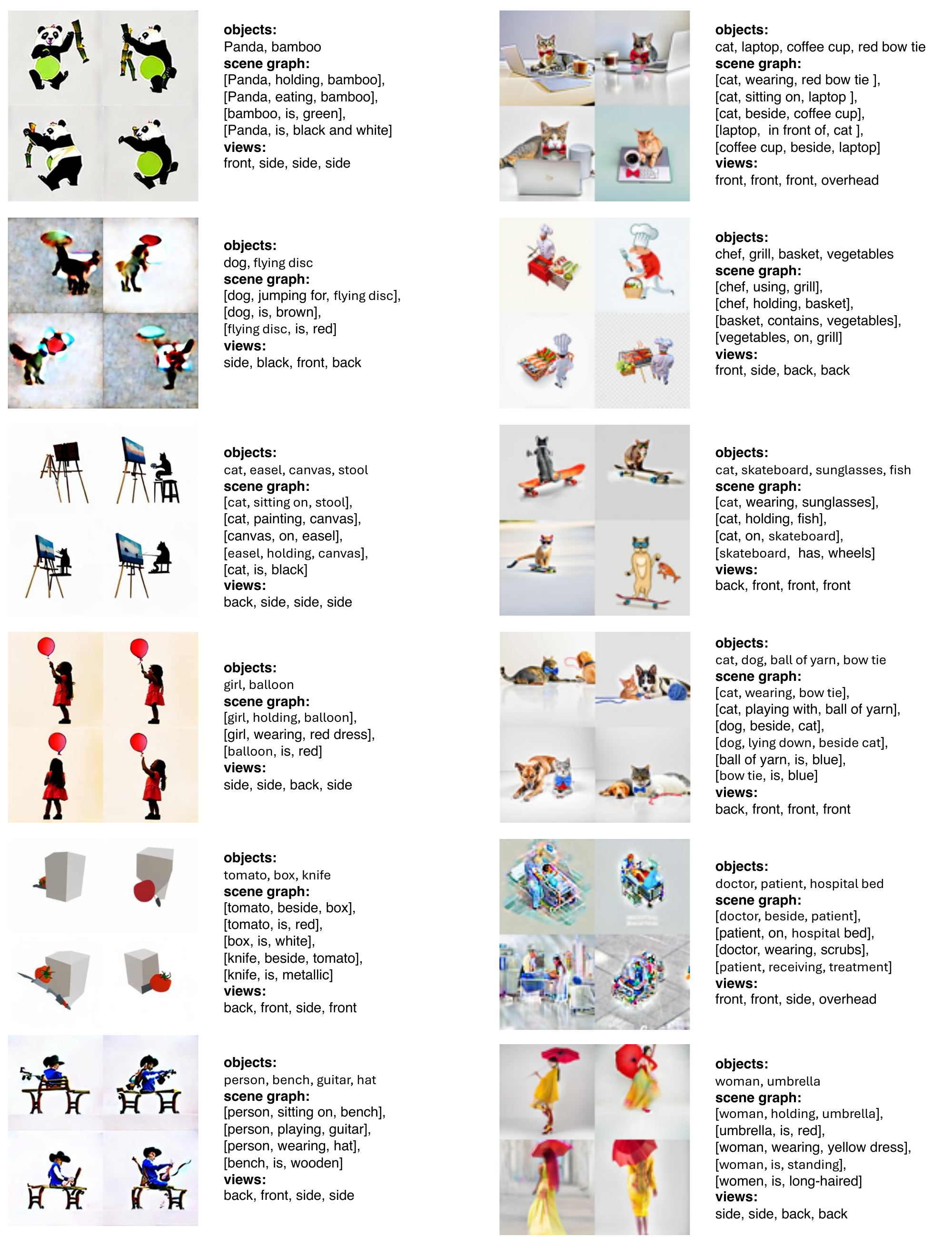}
   \vspace{-3mm}
   \caption{\textbf{Visualization of samples in view-aware dataset.}}
\label{fig:supp-data}
\vspace{-2mm}
\end{figure*}

In \cref{fig:supp-data}, we show visualizations of samples with corresponding answers to three guidance tasks. Since all the meaningful answers can be automatically generated for sampled grid images, we can easily scale up the dataset and enable efficient training of 3DLLaVA-CRITIC.

\begin{figure*}[t]
\hsize=\textwidth 
  \centering
  \includegraphics[width=0.98\textwidth]{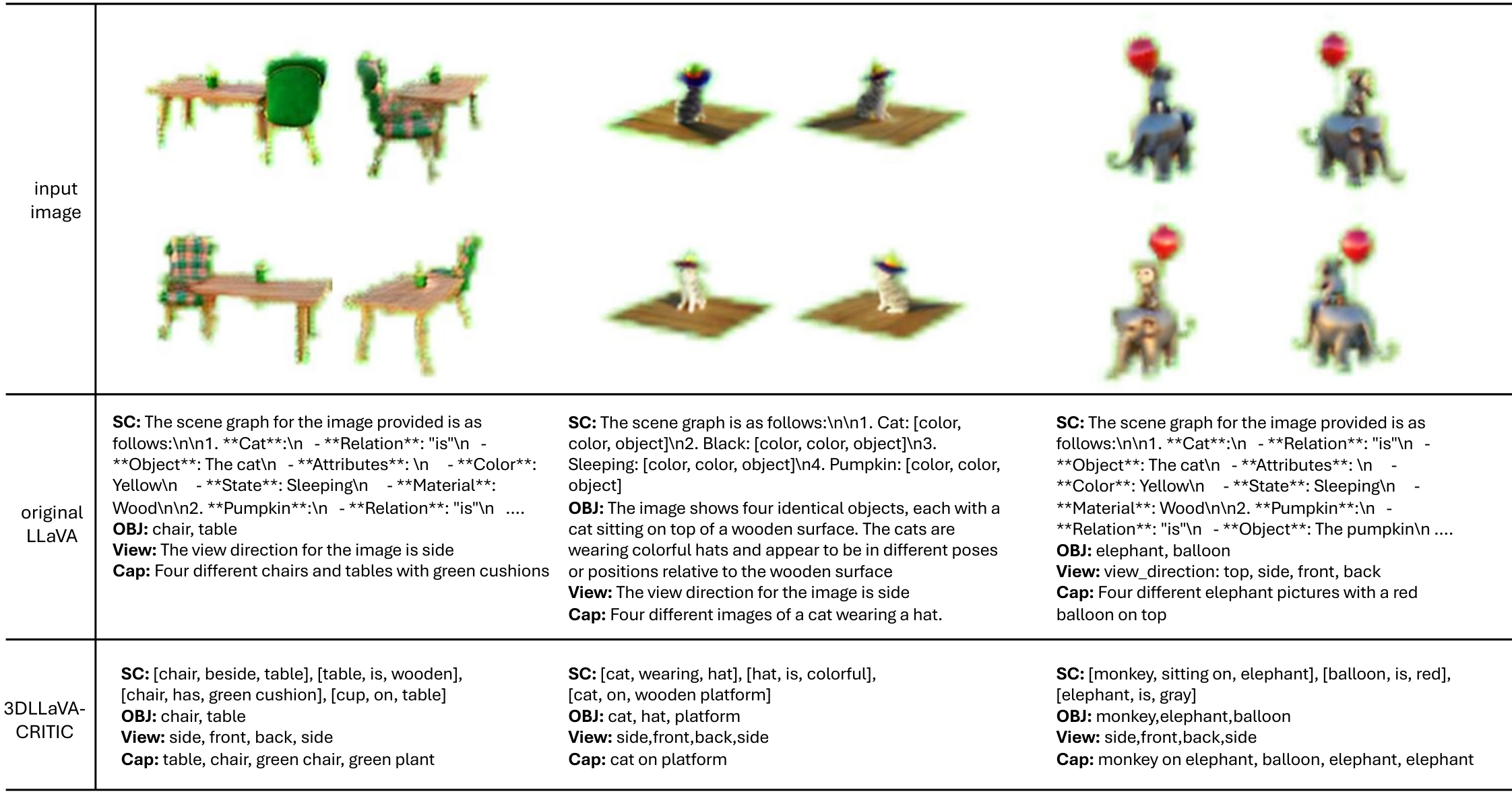}
   \vspace{-3mm}
   \caption{\textbf{Qualitative results of our 3DLLaVA-CRITIC compared to original LLaVA.} SC: generated scene graph; OBJ: predicted multi-label classification; View: view classification; Cap: generated caption.}
\label{fig:supp-llava}
\vspace{-2mm}
\end{figure*}

\subsection{Details of CoherenDream Framework}
We implement CoherenDream by building upon ThreeStudio~\cite{threestudio2023} and the multi-view diffusion model from MVDream~\cite{shi2023mvdream} for text-to-3D generation. The training consists of 5,000 steps, utilizing the AdamW optimizer with a learning rate of 0.01. This process is conducted on a single Nvidia Tesla A800 GPU, with a batch size of 8 and resolution 64 for the initial 3,000 iterations and 4 for the subsequent 2,000 iterations with resolution 256.
To enhance training quality, we incorporate established techniques such as time annealing and resolution scaling from existing open-source implementations. Additionally, we introduce a novel mechanism, the LLM-Layout initialization strategy. Specifically, we warm up the density network of NeRF during the first 600 training iterations using LLM-Layout. In this phase, the loss weight associated with LLM-Layout decreases from 1 to 1e-3 for 600 steps, after which the loss is disregarded beyond 1,000 steps.
Overall, the time consumption for training CoherenDream is approximately 60 minutes, which is comparable to the training time of MVDream but enhances the text alignment.

\input{tab/task_prompt}

\paragraph{Guidance Task Combo. }\label{subsec:supp-guidance}
In this work, we guide the optimization of Textual Coherent Score Distillation (TCSD) using the 3DLLaVA-CRITIC across three guidance tasks: scene graph generation, view classification, and multi-label image classification. As discussed in the main body, scene graph generation provides essential global semantic information, while view classification enhances geometric accuracy. Additionally, multi-label image classification serves to identify any missing objects in the scene. To effectively balance these contributions, we set the weights for optimization as follows: $\lambda_{\text{sg}}=1$, $\lambda_{\text{ob}}=0.5$, and $\lambda_{\text{view}}=0.1$. It is important to note that view classification requires encoding the camera pose at each iteration, which can slow the training process. Therefore, to maintain training efficiency, we compute view classification every 20 iterations.

\subsection{Baselines Setup}
All baselines are reproduced by its officially released code and default parameters. To make a fair comparison with our CoherenDream, we equip the same batch size and resolution. For view prompts used in \footnote{https://github.com/iSEE-Laboratory/DreamView}DreamView~\cite{yan2025dreamview}, objects are removed from the corresponding view prompts when there is a significant semantic occlusion relationship. Otherwise, the view prompts are consistent with the global prompt. For example, the global prompt “A chef is chopping vegetables on a wooden cutting board beside a stove” corresponds to the back view prompt “A chef beside a stove.”. 
For \footnote{https://github.com/chanyn/JointDreamer}JointDreamer~\cite{jiang2025jointdreamer}, we adopt a configuration that involves training for 5,000 steps without quality enhancing strategies: geometry fading and CFG switching. This adjustment is aimed at achieving efficient training and paying attention to text-3D alignment.

\section{Prompts Design}\label{sec:supp-prompt}
\begin{figure*}[t]
\hsize=\textwidth 
  \centering
  \includegraphics[width=0.98\textwidth]{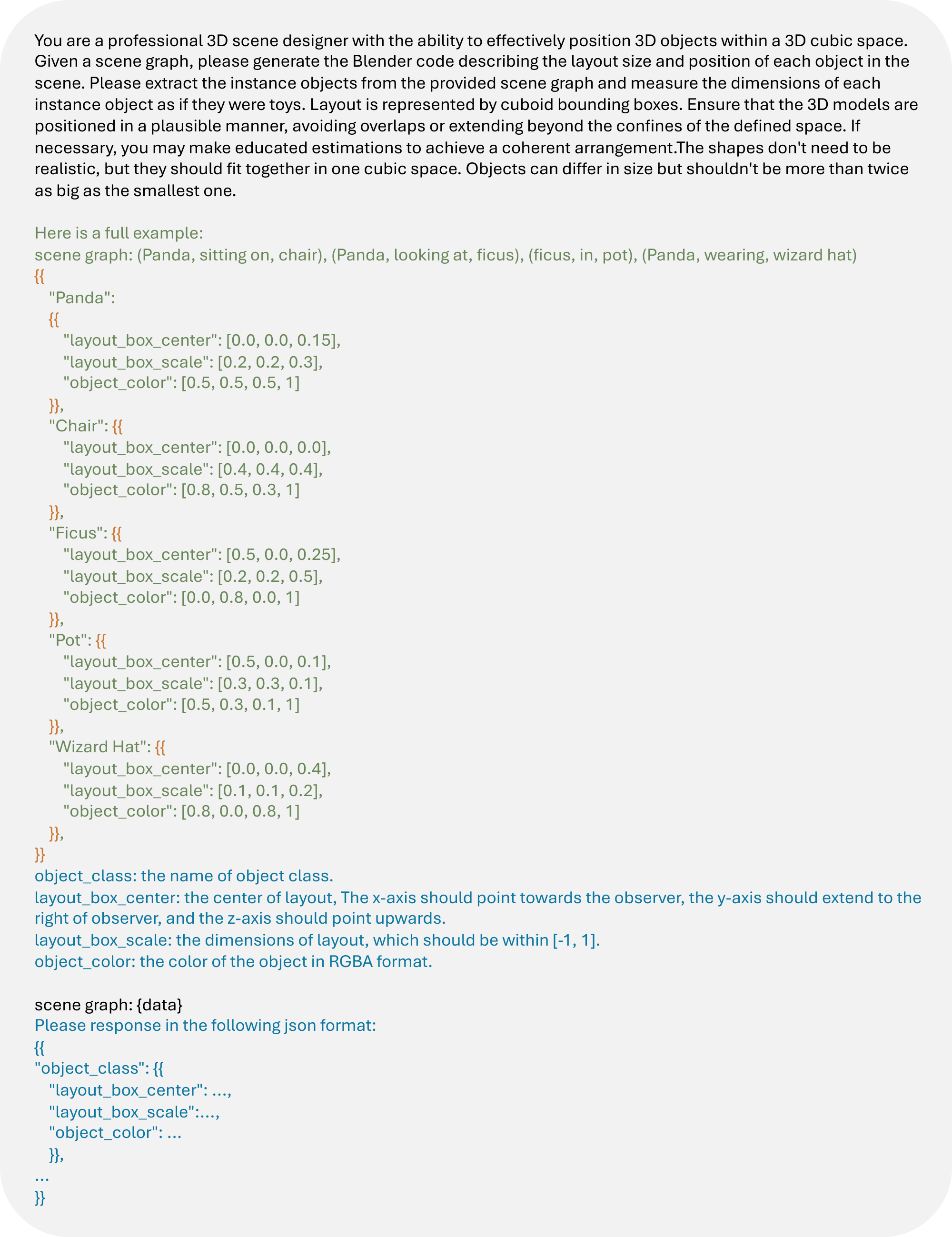}
   \vspace{-3mm}
   \caption{\textbf{Prompt for layout generation.} }
\label{fig:supp-prompt-layout}
\vspace{-2mm}
\end{figure*}

\cref{tab:supp-prompts} presents the prompts utilized for guidance tasks within our Textual Coherent Score Distillation (TCSD) framework, aimed at obtaining semantic feedback from multimodal large language models. Additionally, the layout generation prompt employed in the view-aware data collection pipeline and for the proposed LLM-layout initialization is detailed in \cref{fig:supp-prompt-layout}.

\section{Additional Ablation Studies}
\subsection{Quality of 3DLLaVA-CRITIC}
We present a qualitative comparison between our 3DLLaVA-CRITIC and the original LLaVA~\cite{li2024llava}, as shown in~\cref{fig:supp-llava}. Our observations indicate that the original LLaVA fails to adhere to the required format and content of the instructions, often producing scene graphs that appear to directly replicate the instructions. Furthermore, it does not recognize that grid images may describe the same 3D object. In contrast, after fine-tuning on our view-aware datasets, the Multi-Layer Language Model (MLLM) demonstrates a capacity to generate more accurate scene graphs that align better with instruction requirements.

\begin{figure}[t]
  \centering
  \includegraphics[width=0.5\textwidth]{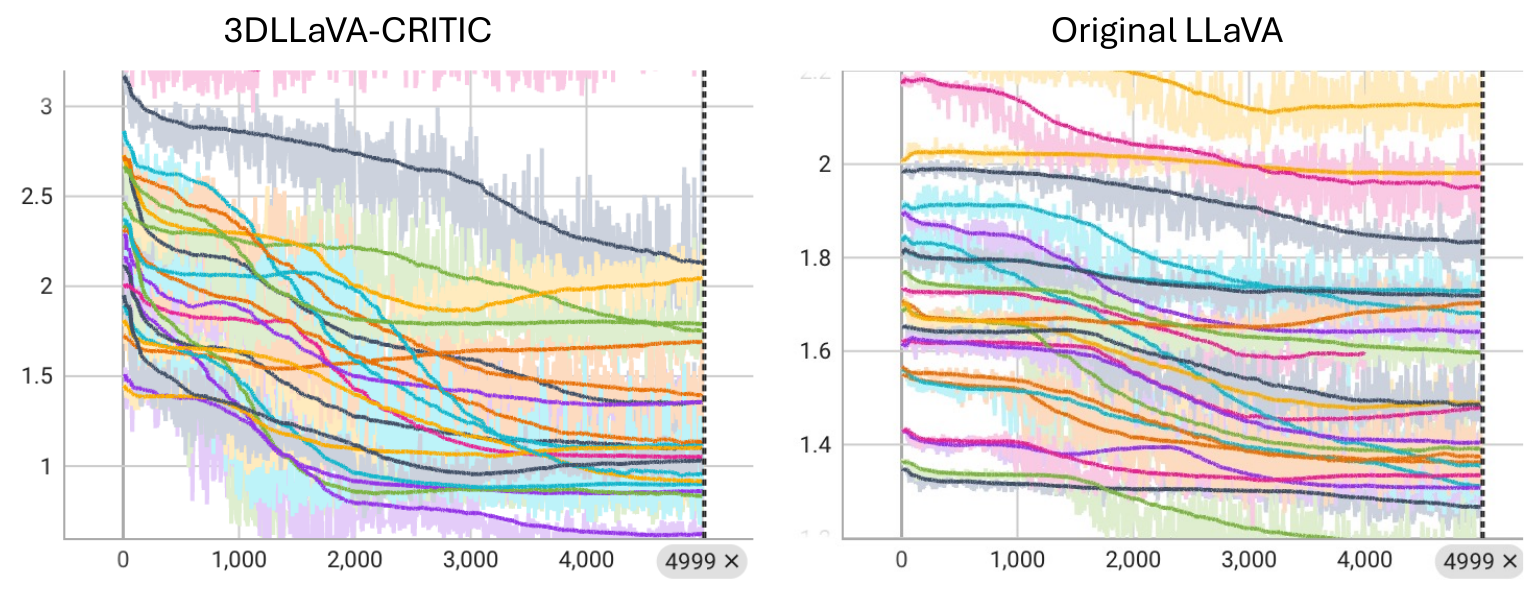}
  \vspace{-3mm}
   \caption{\textbf{Comparison on aggregated loss curve of $L_{\text{MLLM}}$} during score distillation between our 3DLLaVA-CRITIC and original LLaVA. }
   \vspace{-4mm}
\label{fig:supp-lossmllm}
\end{figure}

We also present the loss curve of $L_{\text{MLLM}}$ during score distillation in~\cref{fig:supp-lossmllm}. The results indicate that the loss curve of the original LLaVA is flatter compared to our 3DLLaVA-CRITIC, suggesting that the original LLaVA experiences limited semantic feedback due to the domain gap between the natural image distribution and the sampled image distribution from the diffusion model. In contrast, our 3DLLaVA-CRITIC, which is fine-tuned on view-aware datasets, demonstrates a better understanding of 3D scenarios, thereby enhancing text-to-3D alignment.

\begin{figure*}[t]
\hsize=\textwidth 
  \centering
  \includegraphics[width=0.98\textwidth]{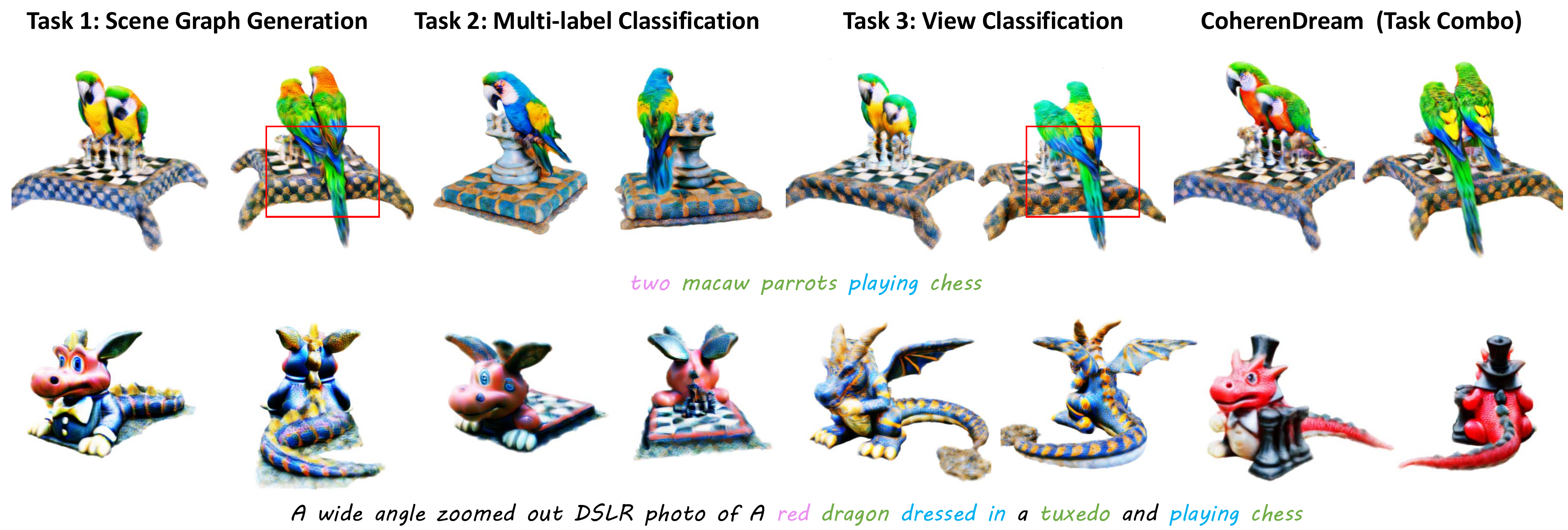}
   \caption{\textbf{Ablation Study on Guidance Tasks with TCSD.} Scene Graph Generation task provides valuable global semantic guidance, resulting in coherent geometric structures, although it may overlook some text-specific details. The Multi-label Classification task ensures correspondence between objects and text; however, it lacks the incorporation of attribution and relational information, leading to unrealistic structures. CoherenDream integrates these three guidance tasks to enhance both text alignment and geometric fidelity in 3D results.}
\label{fig:ablation_guidance}
\end{figure*}

\subsection{Ablation on Guidance Tasks}
To discuss the effectiveness of the decomposed guidance tasks, we conduct studies based on the overall framework of CoherenDream, replacing the combined guidance tasks with individual ones. As shown in~\cref{fig:ablation_guidance}, the scene graph generation task provides global semantic feedback and ensures geometric fidelity. In contrast, the multi-label classification task focuses on the objects present in the image distribution but ignores their attributes and relationships, which can lead to unrealistic placements. The view classification task enhances 3D consistency but it is similar with multi-label classification, ignoring object's attribute. Additionally, view classification requires encoding the camera pose as $\textbf{T}_{vc}$, resulting in slower training speeds. Our proposed CoherenDream integrates these three guidance tasks to enhance both text alignment and geometric fidelity in 3D contents.

We further conduct quantitative ablation studies to evaluate the effectiveness of different guidance tasks on the TIFA v1.0 subset, using TIFA score as the evaluation metric. Specifically, we compare against a variant that uses only captioning as the guidance task, denoted as w/$\textbf{T}_{caption}$. As discussed in Section: Method — 3DLLaVA-CRITIC, we choose scene graph generation as the primary guidance signal because, unlike captions—which often include irrelevant or ambiguous descriptions—scene graphs provide structured supervision that explicitly captures object relationships. The results in Table~\ref{tab:r2} show that removing scene graph generation leads to the most significant performance drop, highlighting its importance. In contrast, removing view classification has a smaller effect on text-image alignment but primarily impacts visualization quality. These findings are consistent with the qualitative results in~\cref{fig:ablation_guidance}.

\begin{table}[h]
\vspace{-5pt}
    \centering
    \resizebox{\linewidth}{!}{
    \begin{tabular}{c|c|ccc}
    \toprule
    CoherenDream & w/ $\textbf{T}_{caption}$ & wo/ $\textbf{T}_{sg}$ & wo/ $\textbf{T}_{ob}$ & wo/ $\textbf{T}_{view}$ \\
    \midrule
    81.4 & 78.4 &  79.1  & 80.5   &  80.9 \\
    \bottomrule         
    \end{tabular}
    }
\vspace{-5pt}
    \caption{Quantity ablation study on guidance tasks.}
    \vspace{-1em}
    \label{tab:r2}
\end{table}

\begin{figure*}[t]
\centering
\includegraphics[width=\linewidth]{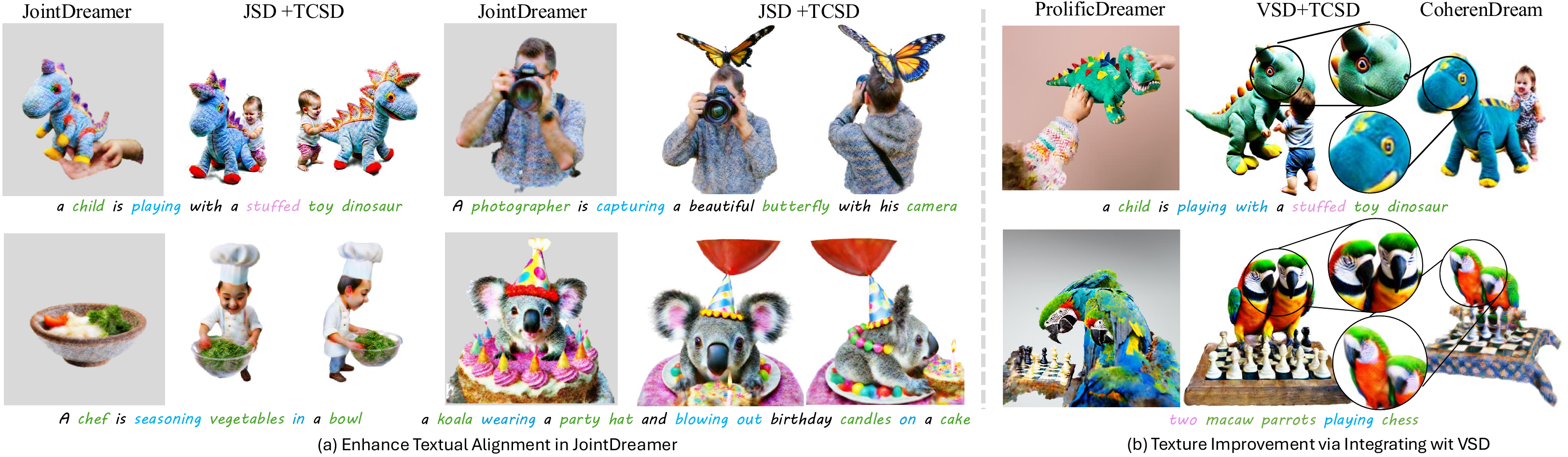}
\vspace{-4mm}
\caption{\textbf{Integrating TCSD with other SDS Variants}.(a) Enhancing prompt alignment: Integrating TCSD with JSD improves textual alignment in JointDreamer, preventing object omission (e.g., child and butterfly) while maintaining original texture style. (b) Improving texture fidelity: Combining with VSD, our method can produce results with higher-quality surface details, as highlighted in the zoom-in regions.}
\label{fig-supp:sds_variants}
\end{figure*}

\begin{figure*}[t]
\hsize=\textwidth 
  \centering
  \includegraphics[width=0.98\textwidth]{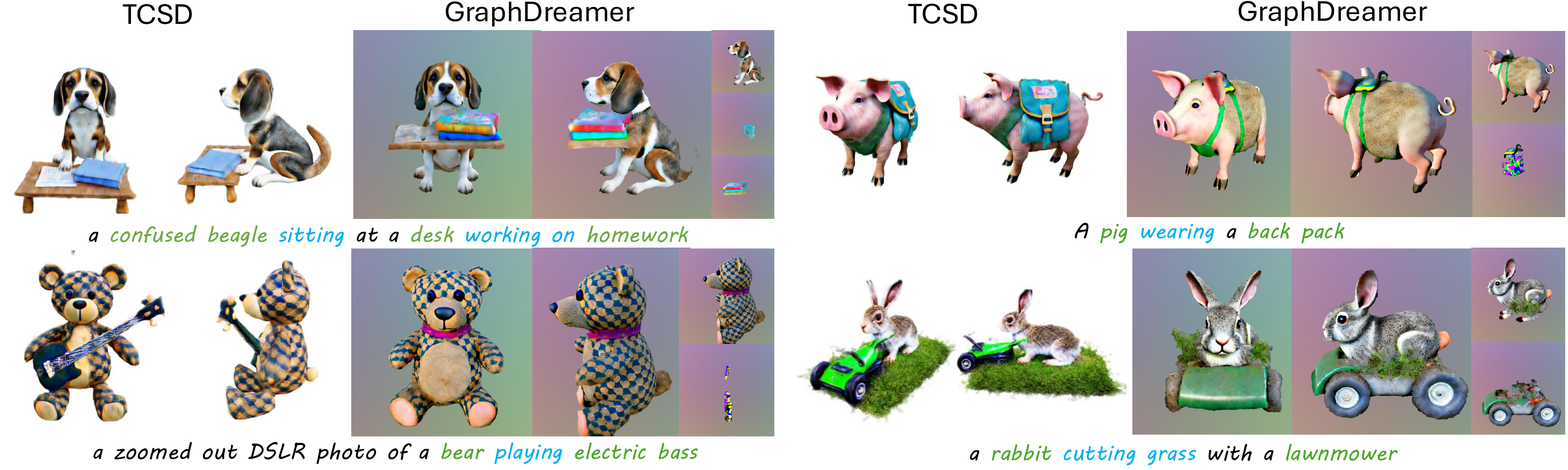}
   \caption{\textbf{More comparison with GraphDreamer,} which encounter unnatural decomposition leading to objects being fully obscured by others. In contrast, TCSD leverages a dynamic MLLM assessor to produce coherent results in a holistic 3D space.}
\label{fig-supp:graphdreamer}
\end{figure*}
\begin{figure}[t]
  \centering
  \includegraphics[width=0.5\textwidth]{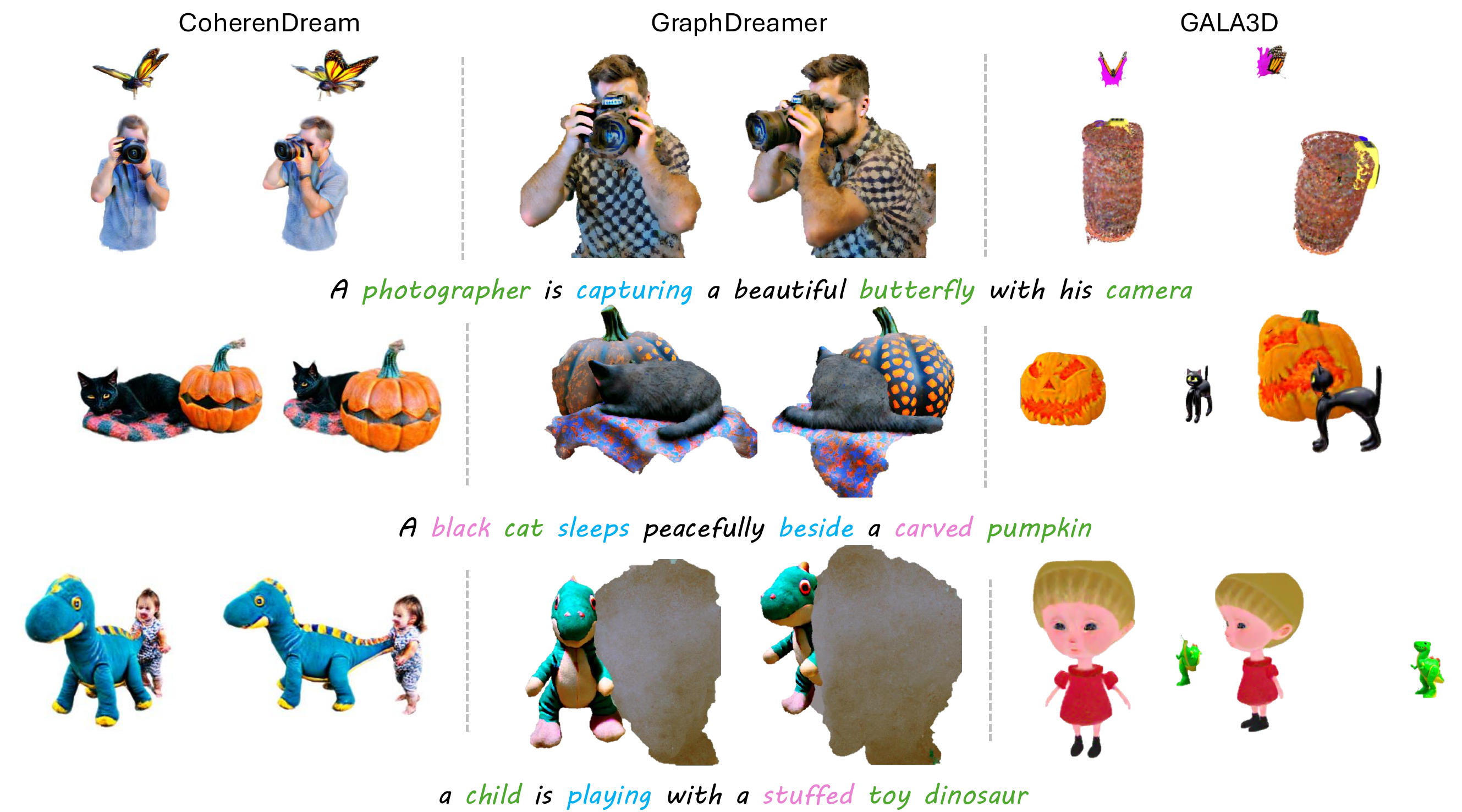}
  \vspace{-3mm}
   \caption{\textbf{Comparison with Compositional Methods.}}
\label{fig-supp:compare_comp}
\end{figure}

\subsection{Integrating TCSD with other SDS Variants}
TCSD primarily enhances prompt alignment and is orthogonal to texture fidelity improvements. We demonstrate its compatibility with other SDS variants by integrating TCSD with JSD~\cite{jiang2025jointdreamer} to address object omission issues in JointDreamer, while preserving its original texture style. Additionally, we combine VSD~\cite{wang2023prolificdreamer} with TCSD to improve texture fidelity, as shown in Figure~\ref{fig-supp:sds_variants}.

\section{Additional Experimental Results}

\subsection{Comparison with Compositional Methods}
We compare CoherenDream with compositional generation methods, including GALA3D\cite{zhou2024gala3d} and GraphDreamer\cite{gao2024graphdreamer}, using the same LLM-layout as initialization and scene graph annotations for all models to ensure fairness. 
GALA3D attempts to use 3D Gaussians representation while dynamically refine layout during generation. However, its non-overlapping constraint leads to unnatural spatial relationships and scale inconsistencies (e.g., excessive spacing between the girl and dinosaur inin~\cref{fig-supp:compare_comp}). 
GraphDreamer advances the field by incorporating automatically parsed scene graphs to guide object interactions. Nevertheless, it inherits the fundamental limitations of compositional approaches, struggling to maintain holistic coherence, particularly in scenes requiring intimate object interactions (e.g., faceless cat in~\cref{fig-supp:compare_comp}).
The failure cases of GraphDreamer are illustrated in~\cref{fig-supp:graphdreamer}, where the model exhibits unnatural object reorganization, often resulting in severe occlusions. In contrast, our TCSD jointly optimizes a holistic 3D representation and leverages informative feedback from MLLMs, leading to user input faithful results.

\input{tab/computation}
\subsection{Computation Overhead Comparison}
We evaluate the computational efficiency of our method by comparing the average training time and peak memory usage with baseline models under the same settings as reported in Table 1 of the main paper. The results are summarized in~\cref{tab:supp-computation}.
Our proposed CoherenDream achieves a training time comparable to MVDream~\cite{shi2023mvdream}, with the main differences stemming from the use of SDS and our proposed Text-Conditioned Score Distillation (TCSD). Importantly, the integration of feedback from the MLLM introduces minimal computational overhead, owing to the efficiency of the LLaVA-OV-0.5B model and the task combination strategy described in \textit{Section Framework of CoherenDream.}
In contrast, GALA3D\cite{zhou2024gala3d}, while benefiting from fast rendering via 3D Gaussian representations, suffers from increased training time due to its compositional optimization, which requires separately optimizing each object. A similar limitation applies to GraphDreamer\cite{gao2024graphdreamer}, whose compositional structure leads to scalability issues as the number of objects increases.

\input{tab/t3bench}
\subsection{Additional Quantitative Comparison on T$^3$Bench}
We perform additional quantitative evaluations on T$^3$Bench~\cite{he2023t} multiple objects set. Following T$^3$Bench~\cite{he2023t}, we employ two complementary metrics: ImageReward~\cite{xu2023imagereward} for assessing 3D asset quality and GPT-4~\cite{wu2024gpt} for text-to-3D alignment verification. As shown in Table~\ref{tab:t3bench}, our CoherenDream achieves comparable mesh and texture quality to the baseline MVDream~\cite{shi2023mvdream}, while demonstrating superior performance in alignment accuracy.

\subsection{Additional Results of CoherenDream} 
\begin{figure*}[t]
\hsize=\textwidth 
  \centering
  \includegraphics[width=1\textwidth]{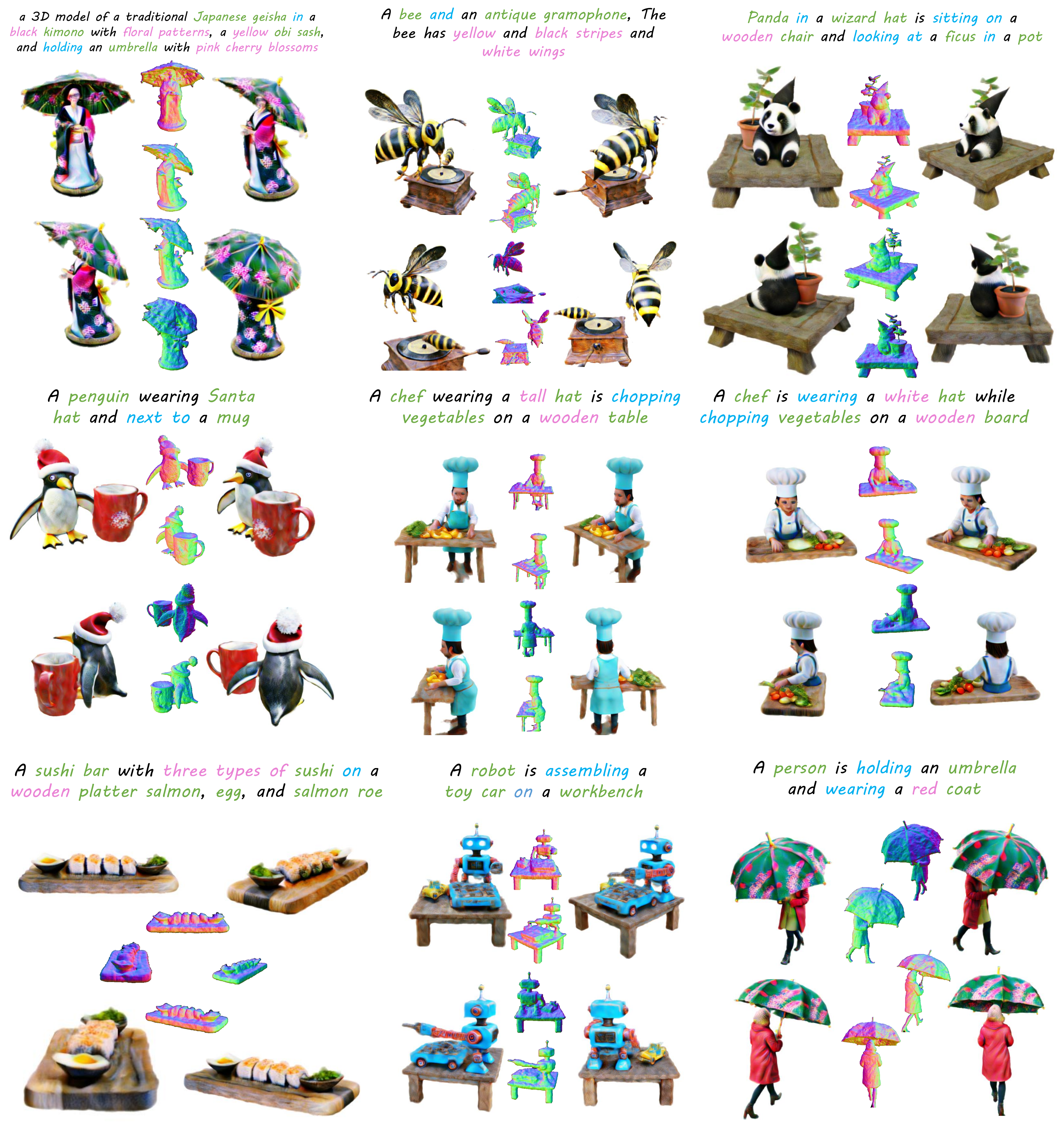}
   \vspace{-3mm}
   \caption{\textbf{More results of CoherenDream.} }
\label{fig:supp-results}
\vspace{-2mm}
\end{figure*}

\begin{figure*}[t]
\hsize=\textwidth 
  \centering
  \includegraphics[width=0.98\textwidth]{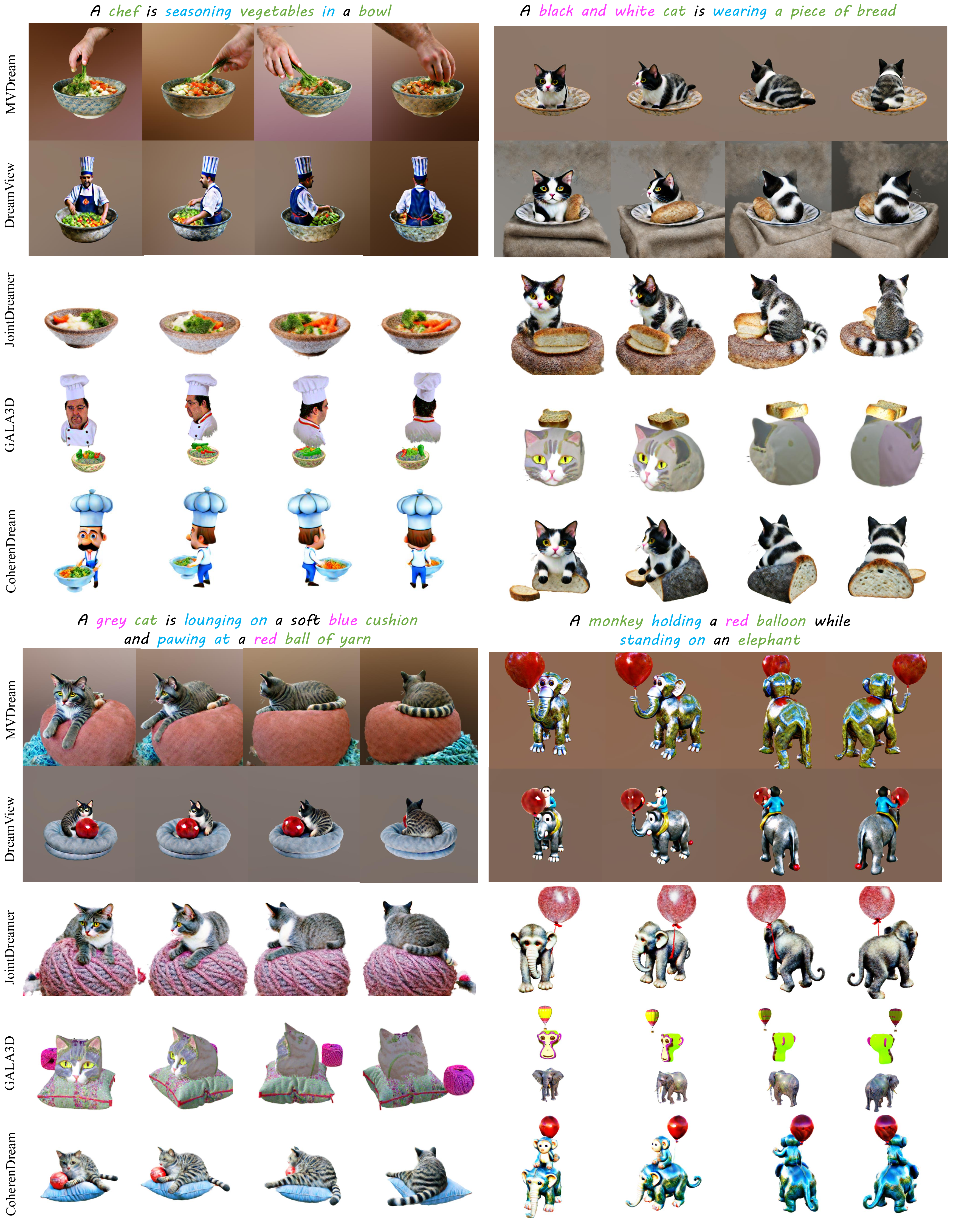}
   \vspace{-3mm}
   \caption{\textbf{More comparison of text-to-3D generation.} }
\label{fig:supp-comp}
\vspace{-2mm}
\end{figure*}

We present more qualitative comparisons of text-to-3D generation with baseline methods as shown in ~\cref{fig:supp-comp}. The results indicate that CoherenDream outperforms current text-to-3D generation methods regarding text consistency. This further validates the effectiveness and generalization of utilizing the text-image understanding and reasoning capabilities of multimodal large language models in 3D generation.
We also provide more images and normal maps from additional generated results in~\cref{fig:supp-results}, demonstrating that the proposed CoherenDream can produce user's input faithful results.

\subsection{Discussion on Failure Cases} 
\begin{figure}[h]
  \centering
  \includegraphics[width=0.45\textwidth]{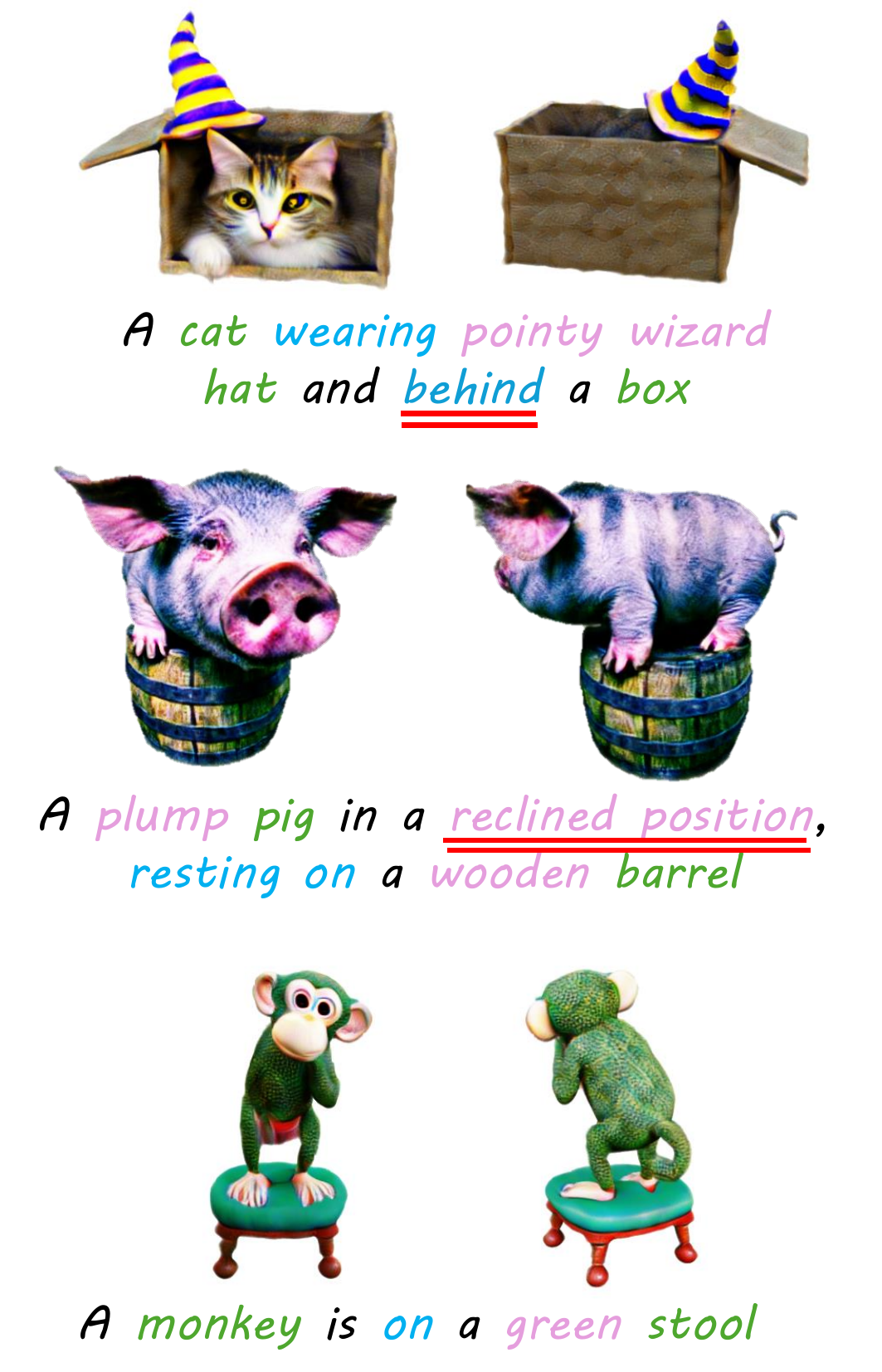}
   \vspace{-3mm}
   \caption{\textbf{Failure cases of CoherenDream.} }
\label{fig:supp-failurecases}
\vspace{-2mm}
\end{figure}

We present some failure cases in Figure~\ref{fig:supp-failurecases}. While CoherenDream demonstrates strong holistic 3D generation and captures complex object interactions, certain fine-grained attributes may be missed. For instance, the ``reclined position" of the pig is not faithfully rendered. Additionally, in the absence of explicit constraints, attribute leakage may occur. In the example ``a monkey is on a green stool" the ``green" attribute incorrectly influences the appearance of the monkey, reflecting the model’s tendency to hallucinate when prompt supervision is weak.

%% file: tab/task_prompt.tex
\begin{table*}[t]
  \centering
  \begin{tabular}{p{3cm}|p{12cm}}
    \toprule
    \centering \textbf{guidance task} & \centering \textbf{prompt} \tabularnewline
    \midrule
    \multirow{12}[0]{*}{Scene Graph} & \textless image\textgreater \textbackslash{}nYou are a proficient 3D scene graph generator. Given a grid image, please generate a scene graph for me based on following requirements.\\
    & Here is a full example:\\
    & scene\_graph: [cat, beside, pumpkin], [cat, is, black], [cat, is, sleeping], [pumpkin, is, carved], [pumpkin, is, one].\\
    & Each scene graph is a collection of triplets in the form of [subject, relation, object] or [object, is, attribute]. Elements of the scene graph include:\\
    & - **Objects**: Primary items in the scene.\\
    & - **Attributes**: Appearances such as color (e.g., "yellow"), state (e.g., "sleeping" or "standing"), material (e.g., "wooden"), and quantity (each category has fewer than 3 instances).\\
    & - **Relations**: These can be\\
    &     \hspace{0.5cm}- **Actions**: like "jumping over" or "driving on".\\
    &     \hspace{0.5cm}- **Spatial Relations**: like "next to", "besides", "under", or "on top of".\\
    &     \hspace{0.5cm}- **Descriptive Verbs**: like "wearing".\\
    &     \hspace{0.5cm}- **Non-specific Connections**: like "and".\\
    &    scene\_graph: \\
    \midrule
    \multirow{9}[0]{*}{View Classification} & \textless image\textgreater \textbackslash{}nGiven a 2x2 grid image, please predict the view direction for me based on the following requirements.\\
    & Here is a full example:\\
    & view\_direction: side, front, back, side. \\
    & 1. The output order should correspond to left-up, right-up, left-down, right-down of the grid image.\\
    & 2. View direction includes: side, front, back, overhead.\\
    &  \hspace{0.5cm}- side: the side view of the object.\\ 
    &  \hspace{0.5cm}- front: the front view of the object, usually including the face of the object.\\
    &  \hspace{0.5cm}- back: the back view of the object. In this view, the face should not be included.\\
    &  \hspace{0.5cm}- overhead: the top view of the object.\\
    & Now please predict the view direction for the following image:\\
    \midrule
    Object Classification & \textless image\textgreater \textbackslash{}nBased on an input image, please extract primary objects for me. \\
    \bottomrule
  \end{tabular}%
  \caption{\textbf{Guidance task prompts used in Textual Coherent Score Distillation,} including scene graph generation (Scene Graph), view classification and multi-label image classification (Object Classification)}
  \label{tab:supp-prompts}%
\end{table*}

%% file: tab/computation.tex
\begin{table}[t]
	\begin{center}
\resizebox{0.47\textwidth}{!}{
    \begin{centering}
    \begin{tabular}{c|cc}
\toprule

Method & Train Time (min.) & Memory (GB)  \tabularnewline\midrule 
MVDream~\cite{shi2023mvdream}& 57  & 30\tabularnewline
DreamView~\cite{yan2025dreamview}& 62 & 70 \tabularnewline
JointDreamer~\cite{jiang2025jointdreamer}& 75 & 40 \tabularnewline
GALA3D~\cite{zhou2024gala3d}& 182 & 20\tabularnewline
GraphDreamer~\cite{gao2024graphdreamer}& 110 & 53 \tabularnewline\bottomrule
CoherenDream& 60 &  55 \tabularnewline\bottomrule
    \end{tabular}
    \end{centering}
}

\end{center}
\vspace{-2mm}
\caption{\textbf{Comparison of computation overhead.}}
\label{tab:supp-computation}
\vspace{-4mm}
\end{table}

%% file: tab/t3bench.tex
\begin{table}[t]
	\begin{center}
\resizebox{0.48\textwidth}{!}{
    \begin{centering}
    \begin{tabular}{c|cc}
\toprule

Method  &Quality $\uparrow $ & ALignment $\uparrow $  \tabularnewline\midrule 
Magic3D~\cite{magic3d22}& 26.6 & 24.8\tabularnewline
RichDreamer~\cite{qiu2024richdreamer}& 34.8 & 22.0\tabularnewline
MVDream~\cite{shi2023mvdream} & 39.0  & 28.5 \tabularnewline
CoherenDream & \textbf{39.4} & \textbf{29.1}  \tabularnewline\bottomrule
\end{tabular}
\end{centering}
}

\end{center}
\vspace{-4mm}
\caption{\textbf{Comparison on T$^3$Bench Multiple Objects Set.}}
\label{tab:t3bench}
\end{table}